\documentclass[journal]{IEEEtran}
\usepackage{booktabs}
\usepackage{wrapfig}
\usepackage{makecell}
\usepackage{float}
\usepackage{amsmath}
\usepackage{tikz}
\usepackage{amsthm}
\usepackage{amsmath}
\usepackage{graphicx}
\usepackage{algorithmic}
\usepackage{subfig}
\usepackage{float}
\usepackage[numbers,sort&compress]{natbib}
\usepackage{algorithm}
\ifCLASSINFOpdf
\else
\fi
\begin{document}
%
\title{A data-driven rutting  depth short-time prediction model with metaheuristic optimization for asphalt pavements based on RIOHTrack}


 \author{Zhuoxuan~Li,
        Iakov~Korovin,
        Xinli~Shi,~\IEEEmembership{Member,~IEEE,}
        Sergey~Gorbachev,        
        Nadezhda~Gorbacheva,
        Wei~Huang,
        and~Jinde~Cao,~\IEEEmembership{Fellow,~IEEE}
\thanks{This work was supported by the Analytical Center for the Government of the Russian Federation (IGK 000000D730321P5Q0002), agreement Nos. 70-2021-00141 (Corresponding author: Jinde Cao.)}   
\thanks{Z. Li is with the School of Mathematics, Southeast University, Nanjing 210096, China. (e-mail: 220191504@seu.edu.cn).}
\thanks{I. Korovin is with the Scientific Research Institute of Multiprocessor Computer Systems, Southern Federal University, Taganrog, 347928, Russia. (e-mail: korovin\_yakov@mail.ru)}
\thanks{X.~Shi is with the School of Cyber Science \& Engineering, Southeast University, Nanjing 210096, China (e-mail:xinli\_shi@seu.edu.cn).}
\thanks{S. Gorbachev is with the Russian Academy of Engineering, Moscow, 125009, Russia. (e-mail:  hanuman1000@mail.ru)}
\thanks{N. Gorbacheva is with the Scientific Research Institute of Multiprocessor Computer Systems, Southern Federal University, Taganrog, 347928, Russia. (e-mail: nadia7@sibmail.com)}
\thanks{W. Huang is with the Intelligent Transportation System Research Center, Southeast University, Nanjing 210096, China (e-mail:hhhwei@126.com).}
\thanks{J. Cao is with the School of Mathematics, Southeast University, Nanjing 210096, China, and with Yonsei Frontier Lab, Yonsei University, Seoul, Korea. (jdcao@seu.edu.cn) }}




\markboth{IEEE/CAA JOURNAL OF AUTOMATICA SINICA,~Vol.~X, No.~X, X~X}%
{Shell \MakeLowercase{\textit{et al.}}: Bare Demo of IEEEtran.cls
for Journals}
%



\maketitle

\begin{abstract}
Rutting of asphalt pavements is a crucial design criterion in various pavement design guides. A good road transportation base can provide security for the transportation of oil and gas in road transportation. This study attempts to develop a robust artificial intelligence model to estimate different asphalt pavements' rutting depth clips, temperature, and load axes as primary characteristics. The experiment data were obtained from 19 asphalt pavements with different crude oil sources on a 2.038 km long full-scale field accelerated pavement test track (RIOHTrack, Road Track Institute) in Tongzhou, Beijing. In addition, this paper also proposes to build complex networks with different pavement rutting depths through complex network methods and the Louvain algorithm for community detection. The most critical structural elements can be selected from different asphalt pavement rutting data, and similar structural elements can be found. An extreme learning machine algorithm with residual correction (RELM) is designed and optimized using an independent adaptive particle swarm algorithm. The experimental results of the proposed method are compared with several classical machine learning algorithms, with predictions of Average Root Mean Squared Error, Average Mean Absolute Error, and Average Mean Absolute Percentage Error for 19 asphalt pavements reaching 1.742, 1.363, and 1.94\% respectively. The experiments demonstrate that the RELM algorithm has an advantage over classical machine learning methods in dealing with non-linear problems in road engineering. Notably, the method ensures the adaptation of the simulated environment to different levels of abstraction through the cognitive analysis of the production environment parameters. It is a promising alternative method that facilitates the rapid assessment of pavement conditions and could be applied in the future to production processes in the oil and gas industry.
\end{abstract}

\begin{IEEEkeywords}
RIOHTrack, Rutting depth, Oil-gas transportation, RELM, Metaheuristic optimization
\end{IEEEkeywords}

%
\IEEEpeerreviewmaketitle

\section{Introduction}
\label{sec1}
%
%
%
%
\IEEEPARstart{V}{igorously} developing the oil and gas industry, energy transportation, and road transportation is one of the essential strategies to fill the sustainable development of energy \cite{gul2019fuzzy}. More importantly, the development of road transportation is one of the foundations for the oil and gas industry. The construction of long-life asphalt pavement is conducive to developing road transportation in the oil and gas industry and ensures the flexibility of oil and gas in road transportation \cite{mir2020impact}. The overall structure of long-life asphalt pavements design is designed for a service life of 40-50 years. One of their main characteristics is that the surface and intermediate layers resist rutting \cite{zheng2020technical}.

Structural damage to the pavement has always affected the service life of roads. Rutting is one of the significant defects of asphalt pavements, affecting roadway travel's quality and safety. This deformation is caused by permanent deformation and shear damage of the asphalt mixture under repeated heavy loads during warm or wet seasons \cite{majidifard2021deep}. This phenomenon is particularly prevalent on heavily traveled routes. The maintenance of asphalt pavements has now increased considerably due to traffic loads and environmental influences, with rutting being one of the most common defects. Maintenance related to rutting requires a great deal of effort from pavement engineers and road agencies. For this reason, developing asphalt pavements with a long service life has become a severe challenge to researchers. Long-life asphalt pavements have become an important trend in pavement development due to structural damage such as fatigue cracking and permanent deformation of conventional asphalt pavements. 

Rutting of asphalt pavements occurs due to accumulated strain, which causes deformation along the wheel tracks of the pavement \cite{rahman2019effect}. The plastic deformation formed along the wheel tracks can cause road users discomfort and reduce the pavement's service life. On the other hand, rutting on the pavement surface reduces the overall strength of the pavement and pavement structure. These pavement diseases can significantly reduce the pavement performance and seriously affect the quality of pavement use and service life \cite{behiry2012fatigue,tayfur2007investigation}. The selection of an appropriate asphalt binder has been reported to improve the resistance of hot mix asphalt (HMA) concrete when subjected to rutting, and fatigue \cite{nagabhushana2013rutting,polacco2015review,porto2019bitumen}.
Furthermore, laboratory testing of asphalt binders for rutting and fatigue requires sophisticated testing equipment that is not readily available in developing countries in tropical regions. However, governments are burdened with providing high-quality pavement structures on a limited budget. In this case, applying predictive modeling will reduce project costs and provide a high-quality pavement structure. As the RIOHTrack experiments are very expensive, the rutting data for each type of road surface are small. Using the data efficiently for high precision rutting depths is a challenge.

Given the above situation, developing a robust artificial intelligence (AI) model for pavement rutting depth prediction is necessary. The research in this paper aims to develop and design a model called IAPSO-RELM and use an equivalent training set for predicting the rutting depths of 19 asphalt pavements with high accuracy. Due to the difficulty of efficiently collecting and using pavement data, complex network methods can build the graphic network and effectively mine data. Through community detection, the most critical structural elements can be selected from different asphalt pavement rut data, and similar structural elements can be found.

In this paper, we calculate the similarity of rutting depth variation of different asphalt pavements by grey correlation, construct a complex network of asphalt pavement rutting depths, and use the Louvain algorithm for community detection of the complex network of asphalt pavements to construct an equivalent training set. An Extreme Learning Machine (ELM) algorithm using residual correction is proposed for the rutting depth prediction task. The independent adaptive particle swarm algorithm is used to optimize the whole model, jumping out of the local optimum using independently adaptive parameters to obtain higher accuracy. Notably, with the current experimental design of RIOHTrack, the method can reduce the frequency of rutting experiments in the field without compromising its estimation accuracy. 

We compare the proposed IAPSO-RELM model with several other methods, including machine learning and deep learning algorithms. The performance of these models is statistically evaluated and validated on RIOHTrack from the Institute of Highway Science, Ministry of Transportation, Beijing \cite{wang2017design}. The contributions are summarized as follows: 
\begin{itemize}
\item This is the first application of the Louvain algorithm to the problem of predicting the depth of rutting in asphalt pavements. A complex network of asphalt pavement rutting is constructed by calculating the similarity of the degree of rutting depth variation of different pavements through grey correlation. The Louvain algorithm is used for community detection to build an equivalent training set for rutting data of asphalt pavement materials of the same category. Compared with the strategy of using a single pavement for training or all kinds of pavements for training, the neural network algorithm improves by discovering the potential characteristics of the asphalt pavement rutting data, proving the effectiveness of the equivalent training set.
\item A residual correction method with random initialization settings is proposed for the ELM algorithm. The residuals from the initial ELM model on the training set are learned by another ELM model, thus improving the stability and generalization ability of the ELM algorithm.
\item Optimizing the RELM algorithm using the IAPSO algorithm can improve accuracy by over 1200\% compared with the unoptimized algorithm. Compared to other widely used PSO algorithms, the RELM algorithm's training accuracy is improved by 20\%.
\item We apply complex network and artificial intelligence algorithms to predict asphalt pavement rutting depth. It can effectively use asphalt pavement rutting data of different materials to improve prediction accuracy, reduce the number of future tests and provide timely warning of asphalt pavement rutting damage.
\end{itemize}

The rest of this paper is organized as follows.  In Section \ref{sec:relatedwork}, we review the related work. In Section \ref{sec:FUNDAMENTAL-CONCEPT}, the fundamental algorithms are introduced for our work. In Section \ref{sec:Experiment and Analysis}, experimental results are discussed and the prediction performance of the proposed algorithm is analyzed.  In Section \ref{sec:Conclusion}, the conclusion of our work is drawn and the future work is prospected. 


\section{Related Work}
\label{sec:relatedwork}

The problem of rutting on asphalt pavement has led many scholars to start their research from a data-driven perspective. Dong et al. \cite{dong2021data} summarize the data methods commonly used in pavement engineering problems and pointed out that in the face of the increasingly accumulated pavement data in the future, choosing the appropriate model and its combination will be critical research. In order to obtain more comprehensive and reliable road data, a full-scale track called RIOHTrack was built in Beijing, China. The experimental site measured and collected a large number of reliable field data, such as rutting depth data and other pavement service data \cite{wang2020key}. Huang et al. \cite{huang2020surface} use the field data collected by RIOHTrack, and the statistical properties of random variables in asphalt pavement are analyzed.  Furthermore, a multiple linear regression algorithm is used to calculate the pavement rutting reliably. Liu et al. \cite{Liu2022evaluation} propose a complex network method to evaluate the performance of RIOHTrack's longitudinal asphalt pavement by collecting data such as rutting and international roughness index. Liu et al. \cite{liu2021rutting} discuss the applicability of different rutting depth prediction models based on empirical mechanics equations for different asphalt pavement structures based on the field data of the RIOHTrack full-scale track. The data-driven method provides a new method for pavement engineering research to analyze pavement performance and effectively improve calculation efficiency and accuracy. However, it needs the support of a large number of measured data.

Recently, a robust artificial intelligence model has been developed continuously and has preliminary application. Advanced learning algorithms have been introduced to model pavement engineering applications. Zhang et al. \cite{zhang2021investigation} develope a field rutting depth prediction model based on the Random Forest algorithm and included four input parameters: Hamburg Wheel Tracking Test (HWTT) rutting depth, pavement age, number of high-temperature hours, and annual average daily truck traffic (AADTT). The model can accurately predict field rutting depth based on a high coefficient of determination and a low standard error of estimation. Huang et al. \cite{huang2020surface} evaluate the statistical characteristics of random variables in asphalt pavements using field data collected from asphalt pavement projects and found a unique formula with fairly good accuracy for evaluating the rutting life of asphalt pavements using multiple linear regression. Uwanuakwa et al. \cite{uwanuakwa2020artificial} propose the Gaussian Process Regression (GPR) algorithm to predict rutting and fatigue parameters for estimating the rutting and fatigue sensitivity of asphalt binders at medium and high pavement temperatures. The results show that the model performs better in estimating rutting parameters than fatigue parameters. Fang et al. \cite{fang2020prediction} denoise the measured rutting depth index (RDI) by wavelet analysis and then applied an autoregressive moving average model (ARMA) from time series analysis to predict RDI. The method that combines wavelet analysis and time series prediction model is referred to as the W-ARMA model in this study. Machine learning algorithms provide a data-driven approach to asphalt pavement rutting depth prediction, which can effectively reduce prediction errors and improve computational efficiency and accuracy but require the support of a large number of actual measurement data. Haddad et al. \cite{dettenborn2020pavement} develop a rutting prediction model with deep neural network (DNN) technology. The predictive capability of the DNN model was compared with the multiple linear regression model fitted using the same dataset. In addition, a generic series of rutting prediction curves corresponding to specific traffic, climate, and performance combinations were developed to make rutting predictions available to all road agencies. Majidifard et al. \cite{majidifard2021deep} use a convolution neural network (CNN) to predict rutting depth in asphalt mixtures using deep learning techniques to simulate the HWTT. Gong et al. \cite{gong2018improving} use a DNN for predicting rutting depth in the MEPDG framework and developed a random forest model to identify the significance of the variables. Yao et al. \cite{yao2019establishment} applied principal component analysis (PCA) to reduce the dimensionality of traffic variables and used neural networks (NN) to develop a rut depth prediction model.

In addition, based on machine learning research in engineering applications \cite{tang2021review,wang2019parameter,karat2022optimal,liu2022minimum,liu2022data,guo2022exponential}, some researchers have proposed fusion models that combine meta-heuristic optimization with machine learning models to achieve better results. Researchers have applied this method to oil-gas systems and pavement engineering. Islam et al. \cite{islam2020holistic} review several nature-inspired metaheuristic optimization techniques and their applications in well location optimization in oil-gas systems. Ng et al. \cite{ng2022well} establish a data-driven model and integrated the PSO algorithm into the training of Support Vector Regression (SVR)  and Multi-layer Perceptron (MLP). The developed model can estimate the oil production of wells in the Volvo oilfield. Qiao et al. \cite{qiao2020nature} combine the MLP neural network technology with five metaheuristic calculation algorithms to accurately estimate the natural gas consumption in the United States. El-Yamany et al. \cite{abdellatif2021optimizing} use a metaheuristic optimizer (water cycle optimization technology) to optimize the proposed hybrid photovoltaic/diesel system in the oil-gas industry. Xu et al. \cite{xu2021structural} propose an artificial neural network optimization method combined with a genetic algorithm. This method is used to optimize the steel-epoxy asphalt pavement structure of the Sutong Yangtze River Bridge. This method has been shown to improve fatigue reliability. Liang et al. \cite{liang2021machine} combine the newly developed multi-objective PSO algorithm with Gaussian process regression, which can effectively solve the design of the asphalt mixture ratio. Metaheuristic algorithms are combined with machine learning algorithms to improve the prediction accuracy for problems related to pavement engineering and oil-gas systems. It also provides an automated and intelligent design method for pavement engineering and oil and gas systems. The method is more effective for pavement engineering studies than traditional methods.

In contrast to the above work, this paper develops a strong artificial intelligence model of high precision for predicting the rut depth of asphalt pavements. Moreover, it has good applicability to the tested asphalt pavement of RIOHTrack. The longitudinal nature of rutting data will provide important insight into the short and long-term service performance of asphalt pavements. The Louvain algorithm used in this paper to construct the equivalent training set also provides a method by which the field data collected from the field observatory can be used effectively.

\section{Principles of Algorithms}
\label{sec:FUNDAMENTAL-CONCEPT}
\subsection{Gray relation analysis}
Gray relation degree is a measure that describes the strength, magnitude, and order of relationships among factors of a  system \cite{ju1982control}. If the shape of the time series curves characterizing these factors is similar, the degree of correlation is high; otherwise, the degree of correlation is low. Since this method neither requires a high sample size nor a typical distribution pattern when analyzing data, and the analysis results are combined with quantitative and qualitative analysis, it is widely used. It has achieved remarkable success in many fields such as economy, society, agriculture, transportation, and mining.

Let $X_i$ be a system factor and its observed data at time $k$ is $x_i(k)$, then the sequence of factor $X_i$ is $X_i= (x_i(1),x_i(2),...,x_i(n))$. $X_0$ is the reference sequence, $X_1$ is the comparison sequence. The sequences $X_0$ and $X_1$ are normalized, in order to eliminate the magnitudes, by the formula: 
\begin{equation}
    \hat{x}_i(k) = \frac{x_i(k)-min(X_i)}{max(X_i)-min(X_i)}.
\end{equation}
Calculate the relation coefficient between the reference series $\hat{X}_0$ and $\hat{X}_1$ by the formula:
\begin{equation}
   r_{01}(k)=\frac{\mathop{min}\limits_{k}\Delta_1(k)+\rho \mathop{max}\limits_{k}\Delta_1(k)}{\Delta_1(k)+\rho \mathop{max}\limits_{k}\Delta_1(k)},
\end{equation}
where $\Delta_1(k)=\lvert \hat{X}_0(k)-\hat{X}_1(k) \rvert$, $\mathop{min}\limits_{k}\Delta_1(k)$ is the minimum value of $\Delta_1(k)$, $\mathop{max}\limits_{k}\Delta_1(k)$ is the maximum value of $\Delta_1(k)$, and $\rho$ is the discriminant coefficient. The coefficient $\rho$ is artificially given to attenuate the effect of distortion of too large the maximum absolute difference to improve the significance of the difference between the correlation coefficients and the artificially given coefficient, whose values range from $[0.1,1.0]$. In the present study, the practice of most scholars is taken as $\rho = 0.5$ \cite{vallee2008grey,ju1982control,hu2018establishing,jiang2018green}. Moreover, the gray relation degree $\delta_{01}$ between the  sequence $X_0$ and the comparison sequence $X_1$ is calculated by the formula:
\begin{equation}
    \delta_{01} = \frac{1}{n}\sum_{k=1}^{n}r_{01}(k).
\end{equation}
\subsection{Louvain Algorithm}
Node clustering of graphs is a powerful tool for extracting network information. The Louvain algorithm is one of the graph clustering algorithms and is a cohesive clustering algorithm based on the modularity theory proposed by Blondel et al. \cite{blondel2008fast}. The modularity is expressed as:
\begin{equation}
\label{Louvain_1}
Q=\left\{
\begin{aligned}
 & \frac{1}{2m}\sum_{l=1}^{k}\sum_{\alpha \in C , \beta \in C}(A_{\alpha\beta}-\frac{\hat{d}_\alpha \hat{d}_\beta}{2m}), & c_{\alpha}= c_{\beta}, \\
 & 0, & c_{\alpha}\neq c_{\beta}
\end{aligned}
\right.
\end{equation}
where $k$ is the number of detected subgroups, $m$ is the number of edges in the large-scale network, $c_{\alpha}$ and $c_{\beta}$ are the communities to which nodes $d_{\alpha}$ and $d_{\beta}$ belong, respectively, and $\hat{d}_{\alpha}$ and $\hat{d}_{\beta}$ are the degrees of nodes $d_{\alpha}$ and $d_{\beta}$ respectively.
Modularity increment is expressed as:

\begin{align}
\Delta Q &=\left[\frac{\sum i n+2 \hat{d}_{\alpha, i n}}{2 m}-\left(\frac{\sum t o t+\hat{d}_{\alpha}}{2 m}\right)\right]\nonumber \\
&-\left[\frac{\sum i n}{2 m}-\left(\frac{\sum t o t}{2 m}\right)^{2}-\left(\frac{\hat{d}_{\alpha}}{2 m}\right)^{2}\right]
\end{align}

where $\sum in$  is the total degree of all edges in the community $c$, $\sum tot$ is the total degree of all edges to nodes in the community $c$, $\hat{d}_{\alpha, \text { in }}$ is the number of edges from $\hat{d}_{\alpha}$ to all nodes in the community $c$, and $m$ is the total number of edges in the network.

The Louvain algorithm process is divided into two main stages:\\
(1) First, the network is initialized, and each node is considered as its independent community; Then, the whole network is traversed in a particular order, and for each node $i$, the change $\Delta Q$ of the module degree in the community after assigning $i$ to the community of its arbitrary neighbor node $j$ is considered. Specifically, if $\Delta Q>0$, node $i$ is moved to the community that makes the most significant change in $\Delta Q$ in the community of the node that makes the most significant change in $\Delta Q$. Otherwise, it remains unchanged. This process is performed for all nodes in the network iteratively until the increment of modularity no longer changes.\\ 
(2) The graph structure has been reconstructed based on the initial partitioning in the first stage. The node's weight in the new graph is equal to the sum of the weights of the edges inside the community, and the weight of the edge is equal to the sum of the weights of the edges outside the community. The first stage is repeated for the new graph until the modularity of the entire graph does not change again. The Louvain algorithm is shown in Algorithm \ref{alg:Louvain}.
\begin{algorithm}[h] 
\caption{ Louvain Algorithm}
\label{alg:Louvain} 
\begin{algorithmic}[1] 
\REQUIRE
The Initial Graph Structure: $G_0 = (V_0,E_0)$;
\ENSURE 
The Graph-Based Community Segmentation: $G = (V,E)$;
\STATE Initialize the node's attribute data and add the Community attribute as the community identifier of the node;
\label{ code:fram:extract }
\STATE Calculate the total number of nodes in the graph: $N$;
\WHILE{Node $n$ is not the terminating node} 
    \FOR{$i=1$ to $N$}
        \STATE Query the neighboring nodes of node $i$ for their Community attribute. denoted as $C_n$;
        \STATE Calculate the incremental modularity $\Delta Q$ after node $i$ has been moved to $C_n$;
        \STATE Record the neighbourhood community $C_{max}$ that maximises the modularity;
        \IF{$C_i$ in which node $i$ is located is not the same as $C_{max}$}
        \STATE Update the Community property of the node to $C_{max}$;
        \STATE Update the current community internal weights;
        \ENDIF
    \STATE Reconstructing the network structure based on the Community property of the nodes;
    \ENDFOR
\ENDWHILE\\
\RETURN $G = (V,E)$
\end{algorithmic}
\end{algorithm}

\subsection{RELM}
ELM is an algorithm of single hidden layer feedforward neural network (SLFN) \cite{huang2004extreme}, which does not need to adjust the training process compared with traditional algorithms such as BP and RNN, and only needs to obtain the optimal solution based on the connection weights and thresholds of the input and hidden layers and setting the number of neurons in the hidden layer. This algorithm has the advantages of simple parameter setting, fast learning speed, and high robustness \cite{huang2006extreme}.

\begin{figure}[H]
	\centering
	\includegraphics[scale=0.53]{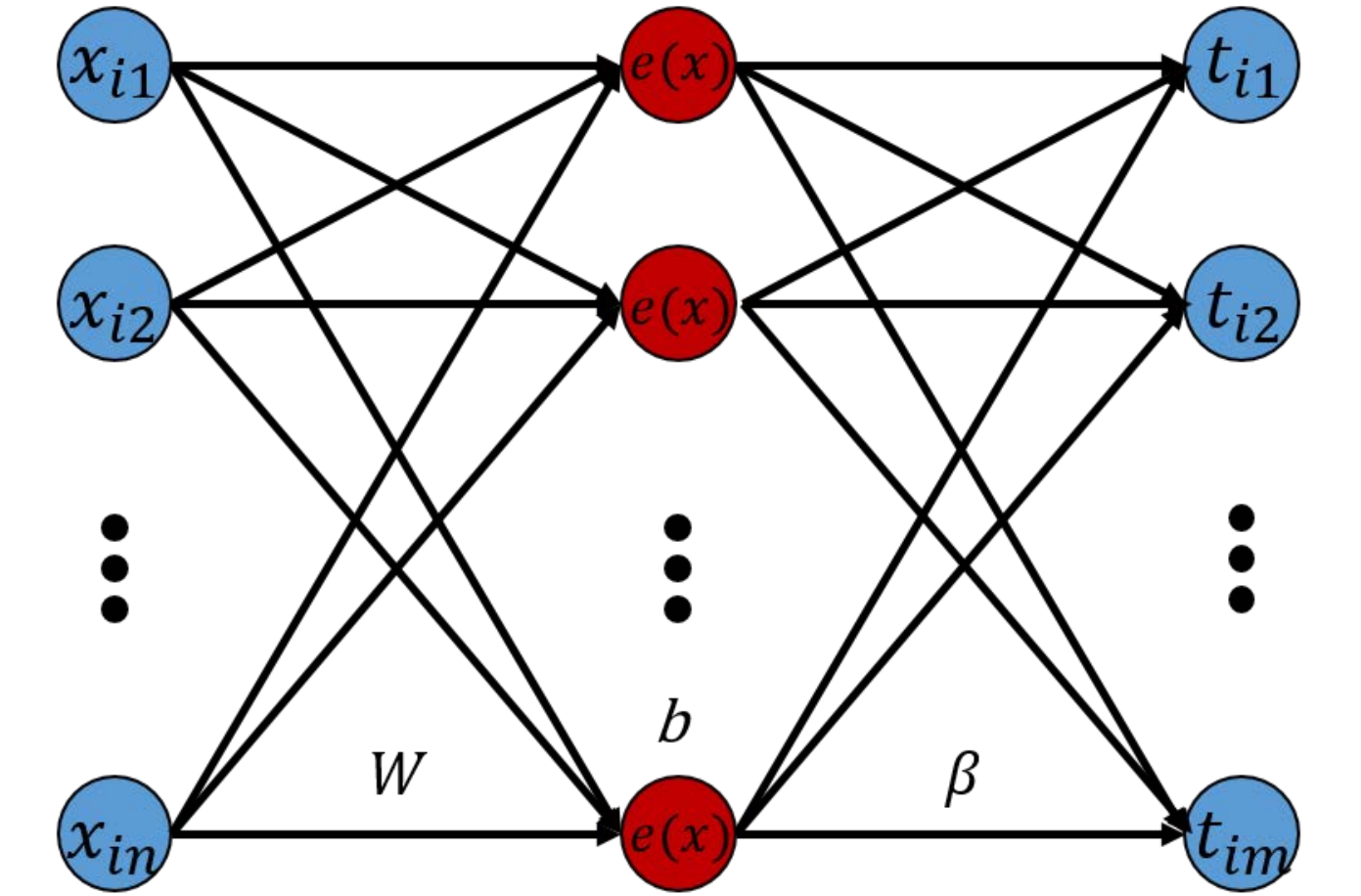}
	\caption{The structure of ELM. }
	\label{fig_ELM}
	\label{fig:visual_smap_gt}
\end{figure}

Assuming that there are $N$ sets of training samples $(X_j, t_j)$, where $X_j = [x_{j1},x_{j2},...,x_{jn}]\in R^n $, $t_j = [t_{j1},t_{j1},...,t_{jm}]\in R^m$,$j = 1,2,...,N$, the ELM hidden layer contains $L$ neurons, and the output of the ELM is: 
\begin{equation}
 \label{ELM_1}
    f(x) = \sum_{i=1}^L[\beta_ie(W_iX_j+b_i)],
\end{equation}
where $e(\cdot)$ is the excitation function; $W_i = [w_{j1}, w_{j2}, ..., w_{jn}]$ is the connection weight between the $i$-th neuron of the hidden layer and each input layer neuron; $\beta_{i} = [\beta_{j1}, \beta_{j2},...,\beta_{jn}]^T$ is the connection weight between the $i$-th neuron of the hidden layer and each output layer neuron; $b_i$ is the threshold of the $i$-th neuron in the hidden layer. The ELM structure is shown in Fig. \ref{fig_ELM}, where $M$ is the the number of nodes in the output layer.

In order to minimize the output error of the single implied layer to achieve the learning objective, Eq.\ref{ELM_1} can be expressed as:
 \begin{equation}
 \label{ELM_2}
     H\beta = T.
\end{equation}

The ELM model can be expressed as the following optimization problem:
 \begin{equation}
 \label{ELM_3} 
   \beta^\dagger = \mathop{argmin}\limits_{\beta}(\frac{1}{2}\Vert \beta \Vert^2 + \frac{C}{2} \Vert H\beta-T \Vert^2),
\end{equation}
where $\beta$ is the connection weight matrix of the hidden layer and the output layer; $\beta^\dagger$ is the solution of the Eq.\ref{ELM_3}; $T$ is the desired output; $H$ is the output matrix of the hidden layer of the neural network; $C$ is the regularization parameter. By the optimality condition, the optimal solution of the optimization Eq.\ref{ELM_3} is:
\begin{equation}
\label{ELM_4}
\beta^\dagger =\left\{
\begin{aligned}
 & (H^TH+\frac{I}{C})^{-1}H^T T, & N\geq L, \\
 & H^T(H^TH+\frac{I}{C})^{-1} T, & N< L,
\end{aligned}
\right.
\end{equation}
where $I$ is the unit matrix.

Since both $W$ and $b$ are randomly given in ELM, the recognition ability of unknown input parameters is weak. This paper proposes an RELM network structure to improve the ELM algorithm. The accuracy and stability are improved by another ELM network named residuals ELM, which learns the residuals of the original ELM on the training set. The residuals of the original ELM on the training set $T_{r}$ is: 
\begin{equation}
\label{RELM_1}
T_{r} = T - H_{o}\beta_{o}^\dagger,
\end{equation}
where $H_{o}$ and $\beta_{o}^\dagger$ are parameters of the original ELM. Next, construct the training set
$(X, T_{r})$. Similarly the $\beta_{r}^\dagger$ of the ELM model with learning residuals is found by Eqs.\ref{ELM_2}-\ref{ELM_4}. Ultimately, the RELM prediction result is:
\begin{equation}
\label{RELM_2}
\hat{T} = H_{o}\beta_{o}^\dagger + H_{r}\beta_{r}^\dagger
\end{equation}
where $\hat{T}$ is the RELM prediction result, $H_{r}$ and $\beta_{r}^\dagger$ are parameters of the ELM model with learning residuals. The RELM algorithm is shown in Algorithm \ref{alg:RELM}.
\begin{algorithm}
\caption{ RELM algorithm}
\label{alg:RELM} 
\begin{algorithmic}[1] 
\REQUIRE Training dataset $(X, T)$, Testing dataset  $\widetilde{X}$, original ELM and residuals ELM;
\ENSURE  
RELM prediction result $\hat{T}$;
\STATE Initialize parameters $W_o$, $b_o$, and $C_o$ of original ELM;
\label{ code:fram:extract }
\STATE Calculate $\beta_{o}^\dagger$ by Eq.\ref{ELM_4};
\label{code:fram:trainbase}
\STATE Calculate $T_{r}$ by Eq.\ref{RELM_1};
\label{code:fram:add}
\STATE Initialize parameters $W_r$, $b_r$, and $C_r$ of residuals ELM; 
\STATE Calculate $\beta_{r}^\dagger$ by Eq.\ref{ELM_4};
\label{code:fram:trainbase}
\label{code:fram:classify}
\STATE Calculate $\hat{T}$ by Eq.\ref{RELM_2}\\
\label{code:fram:select}
\RETURN $\hat{T}$; 
\end{algorithmic}
\end{algorithm}

\subsection{Independent Adaptive Particle Swarm Optimization (IAPSO)}
The PSO algorithm is a robust metaheuristic technique proposed by Eberhart and Kennedy (1995) based on the behavior of the particles/social animals, like birds in a swarm \cite{kennedy1995particle}.  To find the optimal solution, the operation process of the PSO algorithm is summarised \cite{shi2001particle}. 

In the initial period of the algorithm, $n$ particles are first randomly initialized in the D-dimensional search space.  After $k$ iterations, the position $(x_{j1}^k,x_{j2}^k,. . . ,x_{jD}^k)$ and speed $(v_{j1}^k,v_{j2}^k,. . . ,v_{jD}^k)$ of the particle $j$ move in the search space at a certain speed. Meanwhile, the individual extreme value $p_{lbest}^k$ and group extreme value $p_{gbest}^k$  are recorded, and then the states of the population are updated. In each iteration, the velocity and position of particle $j$ can be updated as:
\begin{equation}
\label{pso_1}
v_{jd}^{k+1}=\omega v_{jd}^{k}+c_1r(p_{lbest}^k-x_{jd})+c_2R(p_{gbest}^k-x_{jd})
\end{equation}
\begin{equation}
\label{pso_2}
x_{jd}^{k+1}=x_{jd}^{k}+v_{jd}^{k}
\end{equation}
where $v_{jd}^{k}$ is the velocity of the $j$-th particle in the $d$-dimensional of the search space at the $k$-th iteration, $v_{jd}^{k+1}$ is the velocity at the $(k+1)$-th iteration, $x_{jd}^{k}$ is the position of the $j$-th particle in the $d$-dimensional of the search space at the $k$-th iteration, $x_{jd}^{k+1}$ is the position at the $(k+1)$-th iteration, $\omega$ is the inertial weight, $c_1$ and $c_2$ are acceleration factors, $r$ and $R$ are random numbers on [0,1], $p_{lbest}^k$ is the historical optimal position of the $j$-th particle at the $k$-th iteration, $p_{gbest}^k$ is the historical optimal position of the population at the $k$-th iteration. 

In the evolutionary process, as the evolutionary ability among particles, the overall evolutionary ability of the population, and the problems solved are different, these parameter settings are fully considered in this paper so that the parameters can be adaptively adjusted in different situations.

$\bullet$ Evolutionary Capacity of the Particle 

In the evolutionary process, the \textsl{evolutionary capacity of the particle} is defined as the ability of a particle to find a better solution compared to other particles, which is calculated as follows:\\
\begin{equation}
\label{iapso_1}
E_{i}^{t}=\left(p_{lworst}^{t}-f_{i}^{t}\right) /\left(p_{lworst}^{t}-p_{lbest}^{t}\right)
\end{equation}
where $f_{i}^{t}$ is the fitness value of the $i$-th particle in generation $t$; $p_{lworst}^{t}$ is the worst fitness of all particles at the value of generation $t$. Therefore, $E_{i}^{t}$ is the evolutionary ability of particle $i$ in generation $t$.

$\bullet$ Evolutionary Capacity of the Population 

In the evolutionary process, the \textsl{evolutionary capacity of the population} is defined as the ability of all particles in the aggregate to find a better solution than the current optimal solution, given as follows:
\begin{equation}
\label{iapso_2}
E_{g}^{t}=p_{gbest}^{t}-p_{gbest}^{t-1}
\end{equation}
where $E_{g}^{t}$ is the evolutionary capacity of the $t$-th generation of the population.

$\bullet$ Evolution rate

In particle evolution, the \textsl{evolution rate} of a particle is defined as the magnitude of the particle's ability to evolve in a population. The evolutionary rate formula for the evolutionary capacity of a particle in a population is as follows:
\begin{equation}
\label{iapso_3}
R_i^{(t+1)}=1 / \sqrt{(E_{g}^{t}+E_{i}^{t})^{2}+(1-2E_{g}^{t}E_{i}^{t})},
\end{equation}
where $R_i^{(t+1)}$ is the evolution rate of particle $i$ at the $(t+1)$-th generation.

If both the evolutionary capacity of the particle and the evolutionary capacity of the population are strong, and the particle evolution rate is low, then the particle will inherit the evolutionary ability of the previous generation of particles more in the next generation.

In the PSO algorithm, the setting of inertia weights determines the exploration and searchability of the particles, which plays a crucial role in the algorithm's performance. The size of the particle's global and local search functions can be adjusted to achieve a balance. If $\omega$ is large, the particle has a robust global search capability; otherwise, the particle has a robust local search capability. The inertia weight is usually set by the strategy of linearly decreasing in a specific interval with the increase of the number of iterations, which is calculated as follows:
\begin{equation}
\label{iapso_4}
\omega_{i}^{t+1}=(1-R_i^{(t+1)})\omega_{max}+R_i^{(t+1)}\omega_{min}
\end{equation}
where $\omega_{max}$ and $\omega_{min}$ are the upper and lower limits of the inertia weights, respectively.

The learning factors are two parameters that reflect the historical best learning ability of the particle and the historical best learning ability of the population, respectively. In this paper, $c_{1}$ is designed as a decreasing function, and $c_{2}$ is designed as an increasing function to adjust the learning factor of each particle in each generation according to the rate of change of the evolutionary ability of the particles. Compared with other classical learning factors, this setting method can ensure the particles' learning ability at the beginning of the iteration and enhance global searchability. The adaptive learning factors guarantee the social learning of the particles in the later stages of the iteration, which facilitates accurate local search and can also be performed by independent particles. The learning factor is adjusted using different evolutionary rates so that the particles adjust their learning patterns according to their conditions. The adjustment formula for learning factors of the particle $i$ is as follows:
\begin{equation}
\label{iapso_5}
\begin{gathered}
\left\{
             \begin{array}{lr}
         c_{1 i}^{t+1}=c_{1}^{\max }-\frac{c_{1}^{\min } \cdot \sin \left(M_{i}^{t+1} \frac{\pi}{2}\right) \cdot t}{T}  \\

        c_{2 i}^{t+1}=c_{2 }^{\max }+\frac{c_{2}^{\min } \cdot \sin \left(M_{i}^{t+1} \frac{\pi}{2}\right) \cdot t}{T}\\
             \end{array}
\right.
\end{gathered}
\end{equation}
where $c_{1}^{\max}$ and $c_{1}^{\min}$ are the maximum and minimum values of the corresponding learning factors, respectively. $c_{1 i}^{t+1}$ is the learning factor $c_{1}$ of the $i$-th particle in $(t+1)$-th iteration. The particle should have a strong self-learning ability at the beginning of the iteration. In this case, the value of $c_{1}$ is larger and $c_{2}$ is smaller. If the evolution rate of particle $i$ is small, the value of $c_{1}$ of the particle will be a little larger than other particles, and $c_{2}$ will be smaller, which is more conducive to the global search of the particle. As the number of iterations increases, the particle should have a strong social learning ability. At this point, the value of $c_{1}$ becomes smaller, while the value of $c_{2}$ becomes larger. If the evolution rate of particle $i$ is larger, the particle has smaller $c_{1}$ and larger $c_{2}$ than other particles, which is more favorable to the local search of the particle. In this paper, the learning factor of each particle is adaptively adjusted according to the rate of change of the evolutionary ability of the particle.

In general, Mean Square Error (MSE) can be used as the fitness function to evaluate the degree of evolution of particles which is defined as follow:
\vspace{-0.25cm}
\begin{equation}
\label{pso_9}  
E = \frac{1}{M}\sum_{i=1}^M(Y_i-y_i^*)^2
\end{equation}
where $M$ represents the number of samples, $Y_i$ is the actual value of the sample data, and $y_i^*$ is the predicted value of the sample.

\begin{algorithm}[H]
\caption{ IAPSO-RELM Algorithm}
\label{alg:IAPSO-RELM} 
\begin{algorithmic}[1] 
\REQUIRE  
Algorithm iteration number $T$, population size $M$, initial maximum position $x_{max}$, initial maximum velocity $v_{max}$, $w_{\max }$, $w_{\min }$, $c_{1}^{\max}$ , $c_{2}^{\max}$, $c_{1}^{\min }$, $c_{2}^{\min }$, training dataset $data_{train}$ and test dataset $data_{test}$;
\ENSURE  
Optimal position $\textbf{G}$, optimal fitness function value $E$;

\WHILE{$t<T$}
\FOR{ $i \in [1,M]$} 
    \STATE Decode $x_i$ and initialize the RELM algorithm parameters according to the decoding result of $x_i$; 
    \STATE Input training dataset $data_{train}$ for training the RELM algorithm:
    \STATE Evaluate the fitness function value of $x_i$, and update $p_{i}^{t}$, $p_{gbest}^{t}$,  $p_{lworst}^{t}$; 
    \STATE According to Eq.\ref{iapso_1}-\ref{iapso_5} update $E_{i}^{t}$, $E_{g}^{\prime}$, $M_{i}^{t+1}$, $\omega_{i}^{t+1}$, $c_{1 i}^{t+1}$, $c_{2 i}^{t+1}$;
    \STATE According to Eq.\ref{pso_1}-\ref{pso_2}, update the position and the velocity of each particle;
\ENDFOR
\ENDWHILE\\
\RETURN $\textbf{G}$, $E$
\end{algorithmic}
\end{algorithm}

\subsection{IAPSO-RELM}
The RELM algorithm has several main parameters, such as the connection weight between the hidden layer and input layer in the original ELM $W^{(o)}$, the threshold of the hidden layer in the original ELM $\textbf{b}^{(o)}$, the connection weight between the hidden layer and input layer in the residuals ELM $W^{(r)}$, the threshold of hidden layer in the residuals ELM $\textbf{b}^{(r)}$.The values of these parameters have a significant impact on the model's performance. Adjusting the parameters usually depends on empirical judgments and traversal experiments. The traditional methods are not effective and lack a theoretical basis. Therefore, this paper optimizes the parameters based on the IAPSO algorithm with MSE as the fitness function and retains the optimal population solution and the individual optimal solution for each iteration. The interaction of these two types of information will evolve toward the global optimum. Since the PSO may suffer from poor convergence accuracy, each particle in IAPSO has independent parameters, thus exhibiting better global optimization capability than the PSO. The IAPSO-RELM Algorithm is shown in Algorithm \ref{alg:IAPSO-RELM}.

\section{Experiment and Analysis}
\label{sec:Experiment and Analysis}

\begin{figure}[t]
	\centering
	\includegraphics[scale=0.25]{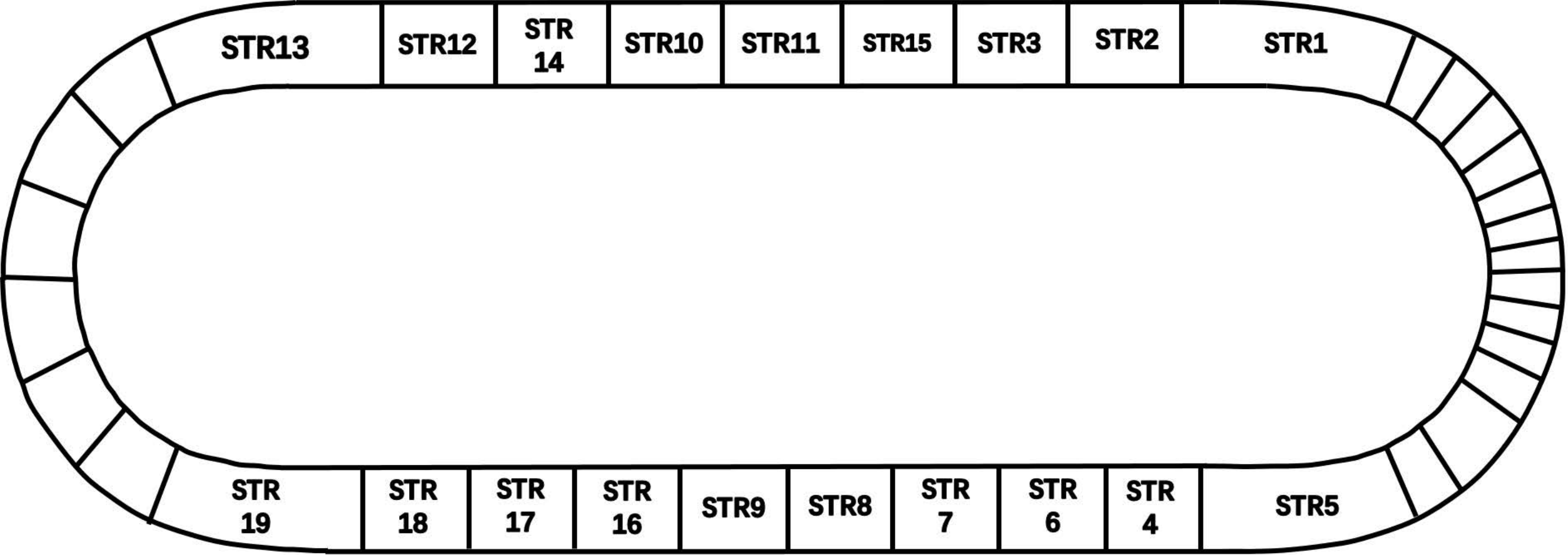}
	\caption{Structure of RIOHTrack.}
	\label{fig_RIOHTrack}
\end{figure}

The measured data of rutting depth come from the full-scale pavement test loop road project of the Ministry of Transport. It is located in the southwest corner of the Beijing Highway Traffic Test Field. It has a 2.038-km-long full-scale field accelerated pavement testing track named as RIOHTrack. The full-scale ring road includes 25 types of asphalt pavement structures. The layout of the pavement structure is shown in Fig.\ref{fig_RIOHTrack}. 

Nineteen main test pavement structures named by STR1-19 were set up in the test ring road to study and compare the long-term performance and evolution of asphalt pavement structures and materials with different combinations of structural rigidity.  The thickness of the asphalt concrete structure layer of these pavement structures includes 12, 18, 24, 28, 36, 48 cm (or 52 cm), which basically covers the thickness of all asphalt concrete structure layers of China's high-grade highways, as well as flexible base thickness of thick asphalt pavement.  From the perspective of the type of base structure, it includes four typical structures: rigid base structure, semi-rigid base structure, flexible base structure, and full-thick asphalt pavement structure \cite{wang2017design}. 

This paper uses measured data from 19 asphalt pavements with full-size pavement structures as the data source. The objective is to make short-time predictions of rut depths for different asphalt pavements. Data from the RIOHTrack experiment for the years 2017-2020 are used for training and data from 2021 for testing. RMSE, MAE, and MAPE are used to evaluate the performance of the algorithms. These metrics are used to reflect the difference between the true and predicted values. Furthermore, they are the most commonly used performance indices in regression tasks. The smaller the index, the more accurate the prediction. These accuracy indices are defined in Table.\ref{table_2}.   
\begin{table}[h]
	\caption{Indicator description}
	\centering
	\label{table_2}
	\begin{tabular}{ll}  	
		\toprule   	
		Index & Formula \\
		\midrule   
		MAE        &$\frac{1}{M}\sum_{m=1}^M  \lvert y_m-\hat y_m \rvert $  \\
        RMSE        &$\sqrt{\frac{1}{M}\sum_{m=1}^M(y_m-\hat y_m)^2}$ \\
        MAPE        &$\frac{100\%}{M}\sum_{m=1}^M  \lvert \frac{y_m-\hat y_m}{y_m} \rvert$ \\
        
		\bottomrule  
	\end{tabular}
\end{table}

\subsection{Model Building}

\begin{figure}[h]
	\centering
	\includegraphics[scale=0.65]{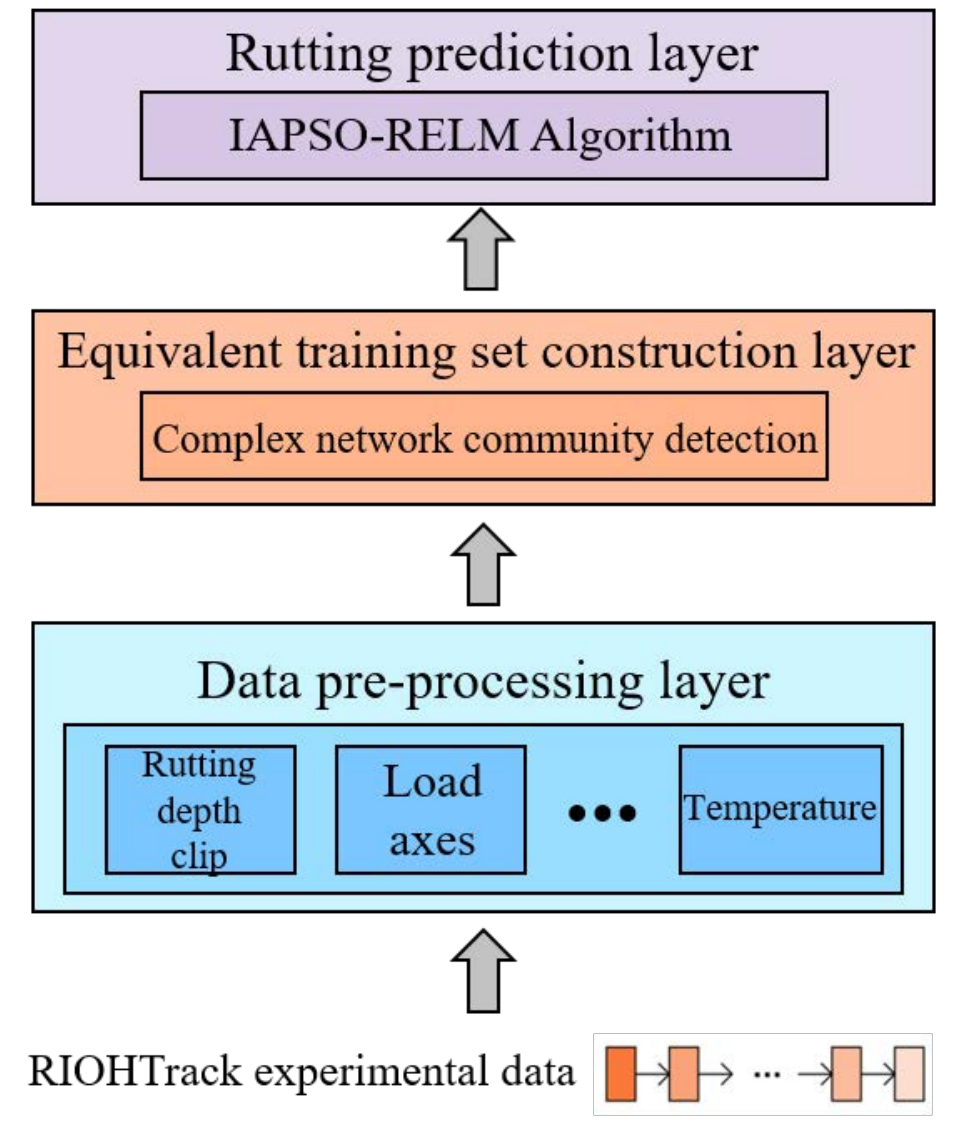}
	\caption{Flow chart of rutting depth prediction.}
	\label{fig_FLOW}
\end{figure}

The framework of the rutting depth prediction model is shown in Fig.\ref{fig_FLOW}, which contains five parts: data processing, constructing complex networks with rutting depth, constructing equivalent training set by community detection, and data fitting.

\begin{figure}[h]
	\centering
	\includegraphics[scale=0.50]{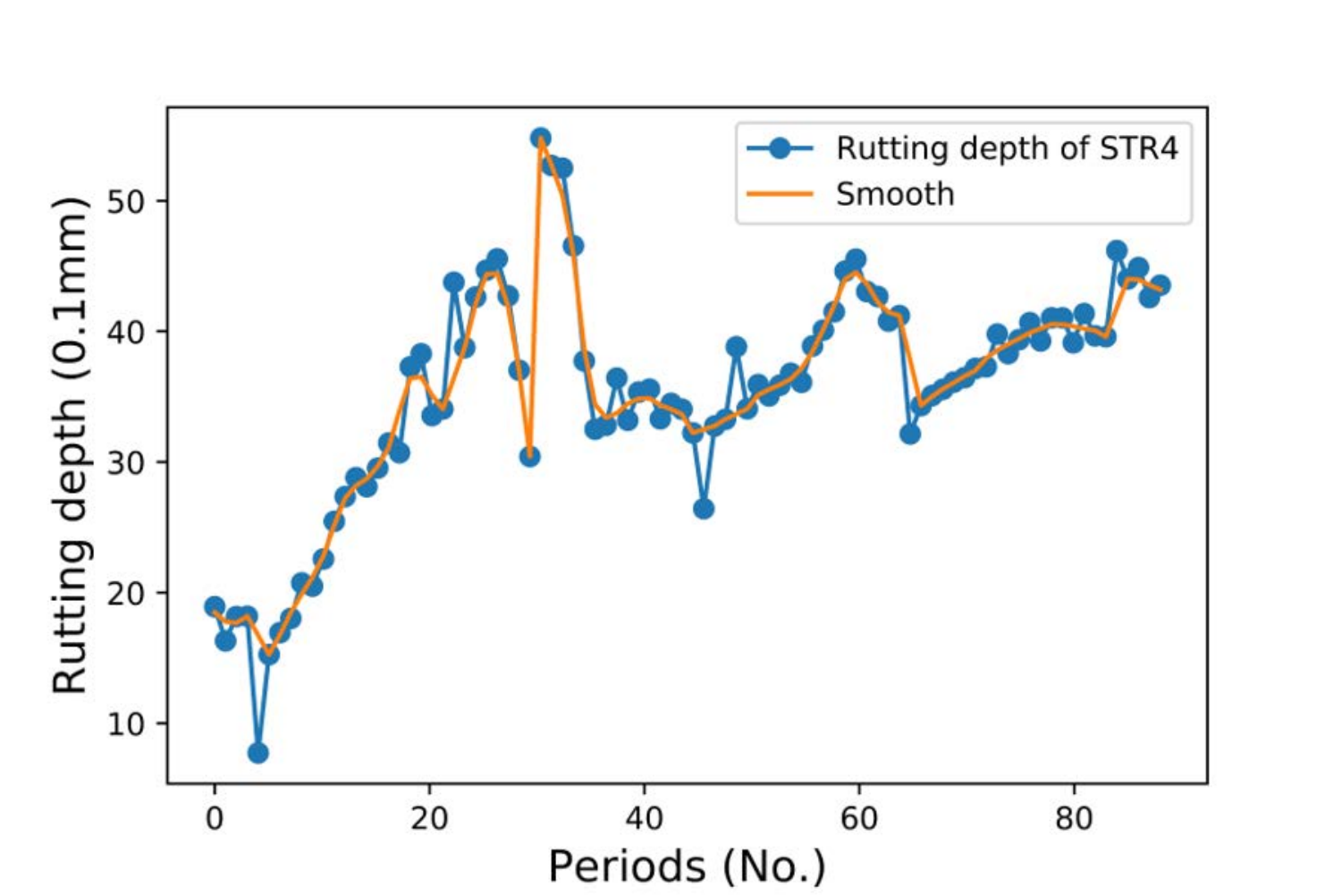}
	\caption{Typical data and smoothed data.} 
	\label{fig_smoothed}
\end{figure}

The initial data of 19 kinds of asphalt pavement mainly include cumulative load axis times, temperature, and rutting depth. New features such as characteristic load axis time, rate of change of load axis time, rate of change of temperature, and the difference in temperature change are generated by calculation. Sliding windows with the time-step of 5 generate rutting depth clips. Take STR4 asphalt pavement as an example. The specific characteristics are shown in Tabel.\ref{table_dataset}. Smoothing of the rutting depth data is used to remove noise from the data. In this paper, the Loess smoothing algorithm \cite{cleveland1988locally} with a span of 0.05 is used, and the smoothing results are shown in the Fig.\ref{fig_smoothed}.
\begin{table}[h]
	\caption{The initial data of STR4 (part)}
	\centering
	\scriptsize
	\label{table_dataset}
	\begin{tabular}{cccccc} 
	    
		\toprule   	
		        \makecell*[c]{Loading\\period}
		      & \makecell*[c]{Accumulated\\load\\axis times}
		      & \makecell*[c]{Rutting \\ depth} 
		      &  . . . 
		      & \makecell*[c]{Temperature \\change \\rate}
		      & \makecell*[c]{Difference in \\temperature \\change}\\
		\midrule   
		N1     & 28396 & 18.934     & . . .  &0              &0\\
        N2     & 104330 &16.32      & . . .  &-0.677515892   &-2.600904262\\
        N3     & 297079 &18.18      & . . .   &-0.287463975  &-0.355874299\\
        N4     & 483685 &15.267       & . . .    &3.090000498  &2.725703262 \\
        N5     & 670604 &56. 47     & . . .  &2.167057124    &7.818325453
\\
        $\vdots$ &$\vdots$ &$\vdots$ &$\vdots$  &$\vdots$  &$\vdots$  \\
        N94   & 41442589 & 43.541& . . .  &-1.054488976
 &4.052485194\\
		\bottomrule  
	\end{tabular}
\end{table}

\subsection{Analysis of community detection of rutting depth in asphalt pavements}
\subsubsection{Complex network construction for asphalt pavements rutting depth}
\begin{figure}[t]
	\centering
	\includegraphics[scale=0.50]{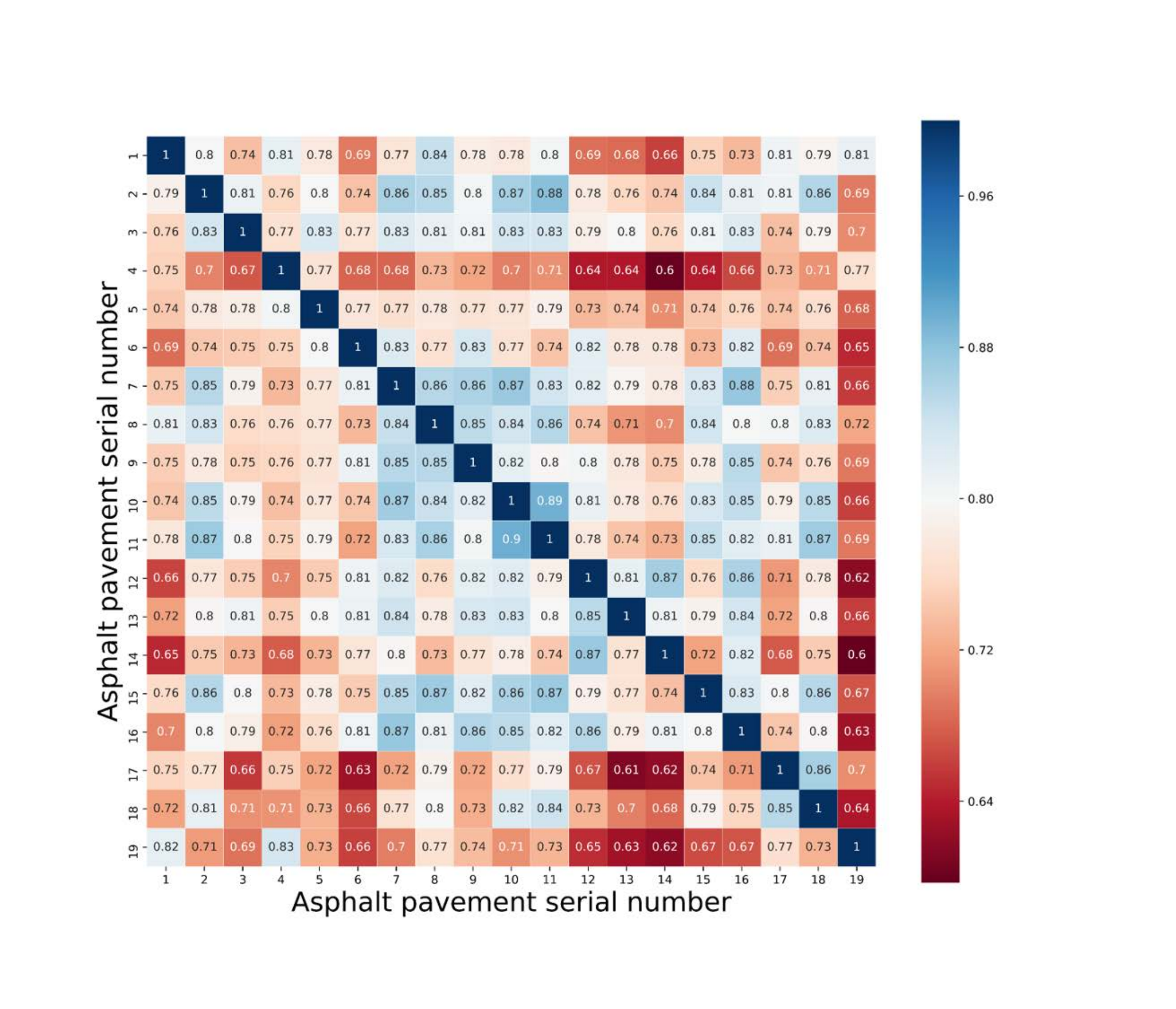}
	\caption{The heatmap of grey correlation of asphalt pavements rutting depth.} 
	\label{fig_heatmap}
\end{figure}

With 19 kinds of asphalt pavements in the RIOHTrack rutting experiment, each containing 88 cycles of sampled rutting data, a complex network can be constructed by abstracting the different asphalt materials as nodes and using grey correlation to calculate the similarity of rutting changes between different asphalt pavements, and the similarity of rutting changes between different asphalt pavements as connected edges of nodes. A heatmap of the connected edge weights of each node is shown in Fig. \ref{fig_heatmap}.

\begin{figure}[h]
	\centering
	\includegraphics[scale=0.39]{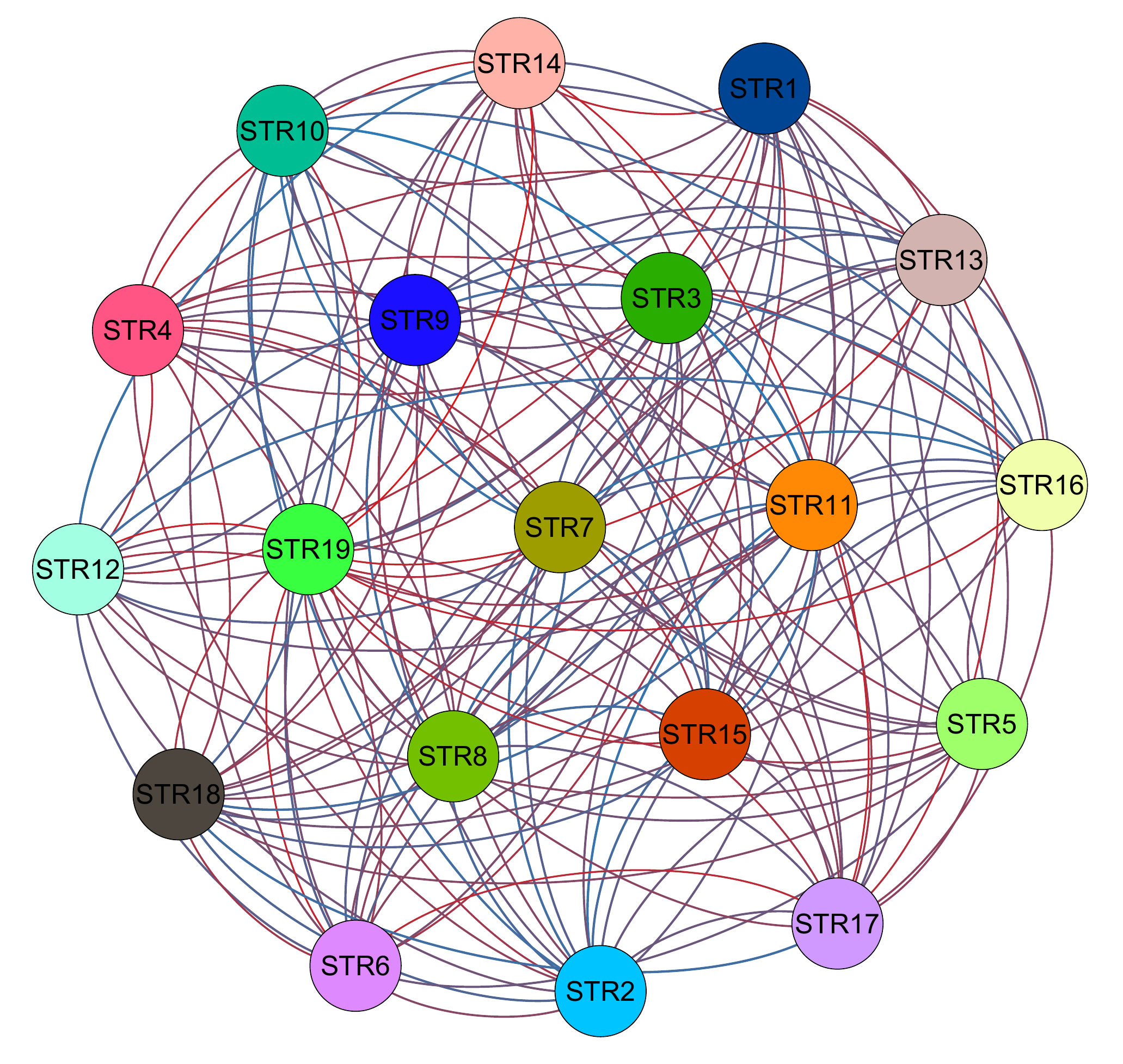}
	\caption{The structure of complex network of asphalt pavements rutting depth. }
	\label{fig_SCN}
\end{figure}
As Fig.\ref{fig_heatmap} shows, the bluer the color, the more similar the rutting variation between the two asphalt pavements. Conversely, they are less similar. The structure of the complex network of asphalt pavements rutting depth is shown in Fig.\ref{fig_SCN}.

\subsubsection{Community detection based on Louvain algorithm}
\begin{figure}[h]
	\centering
	\includegraphics[scale=0.39]{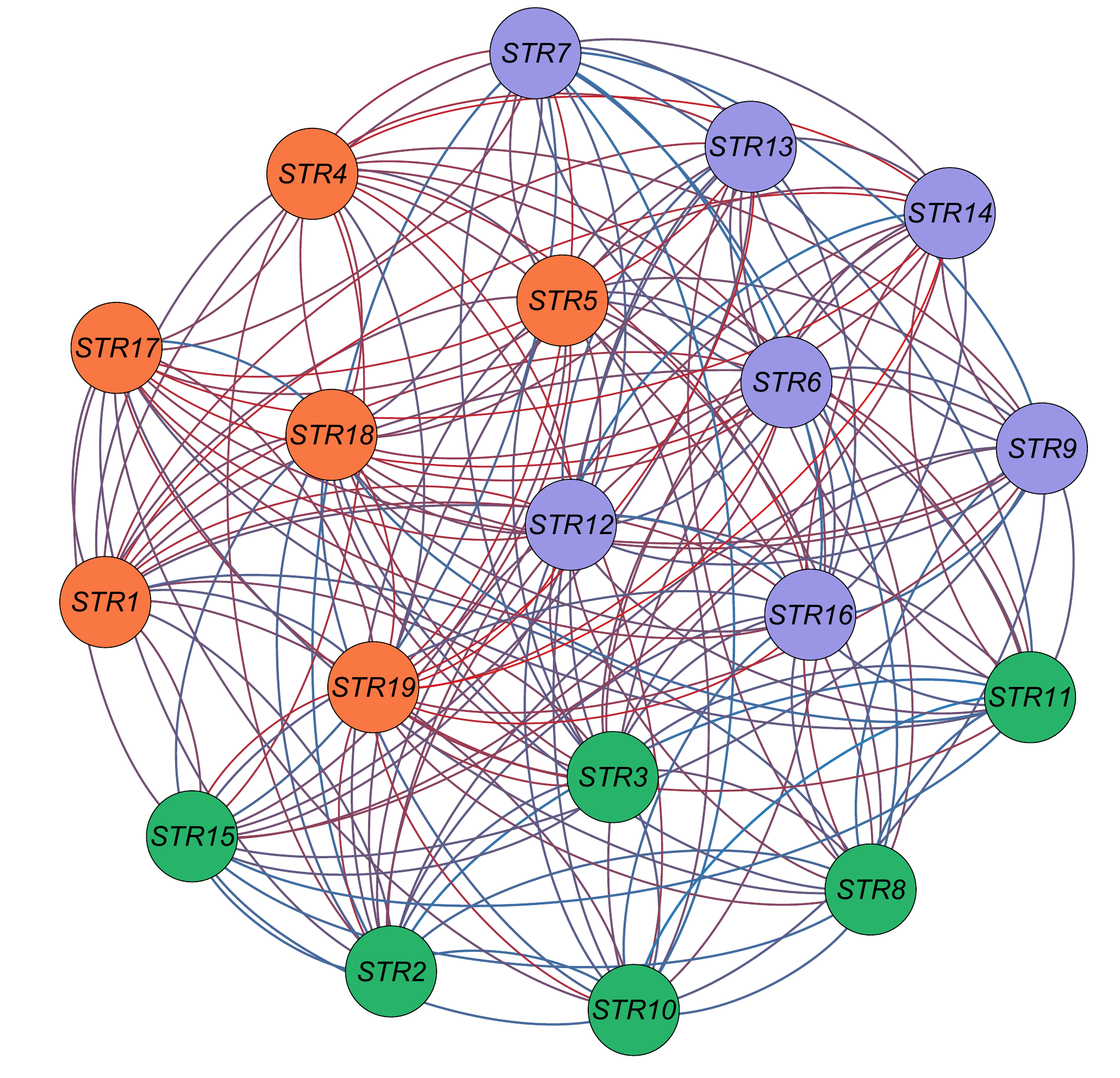}
	\caption{The structure of complex network after community detection. }
	\label{fig_SDSCN}
	\label{fig:visual_smap_gt}
\end{figure}

To categorize asphalt pavements with the same trend of rutting depth variation, the Louvain algorithm is used for community detection of complex networks constructed based on the variation of asphalt pavement rutting depth. The results of the community detection are shown in Table \ref{table_CD} and the structure of the complex network after community detection is shown in Fig. \ref{fig_SDSCN}.   

\begin{table}[h]
    \scriptsize
	\caption{The results of the community detection}
	\centering
	\label{table_CD}
	\begin{tabular}{llll} 
		\toprule   	
		  & Group I & Group II & Group III \\
		\midrule   
		Serial number           & STR1,STR4,STR5,   & STR6,STR7,STR9,   & STR2,STR3,\\
        of asphalt              & STR17,STR18,      & STR12,STR13,      & STR8,STR10,\\
        pavements               & STR19             & STR14,STR16       & STR11,STR15\\
		\bottomrule  
	\end{tabular}
\end{table}
Based on the change in asphalt pavements rutting depth from 2017-2019, STR1, 4, 5,17 18 and 19 are classified in the same group (Group I), while STR6, 7, 9, 16 and 12-14 belong to the same group (Group II) and the rest are considered as another group (Group III).

\begin{figure*}[t]
    \centering
	  \subfloat[GRU with a Single Kind of Asphalt Pavement Data]{
       \includegraphics[width=0.31\linewidth]{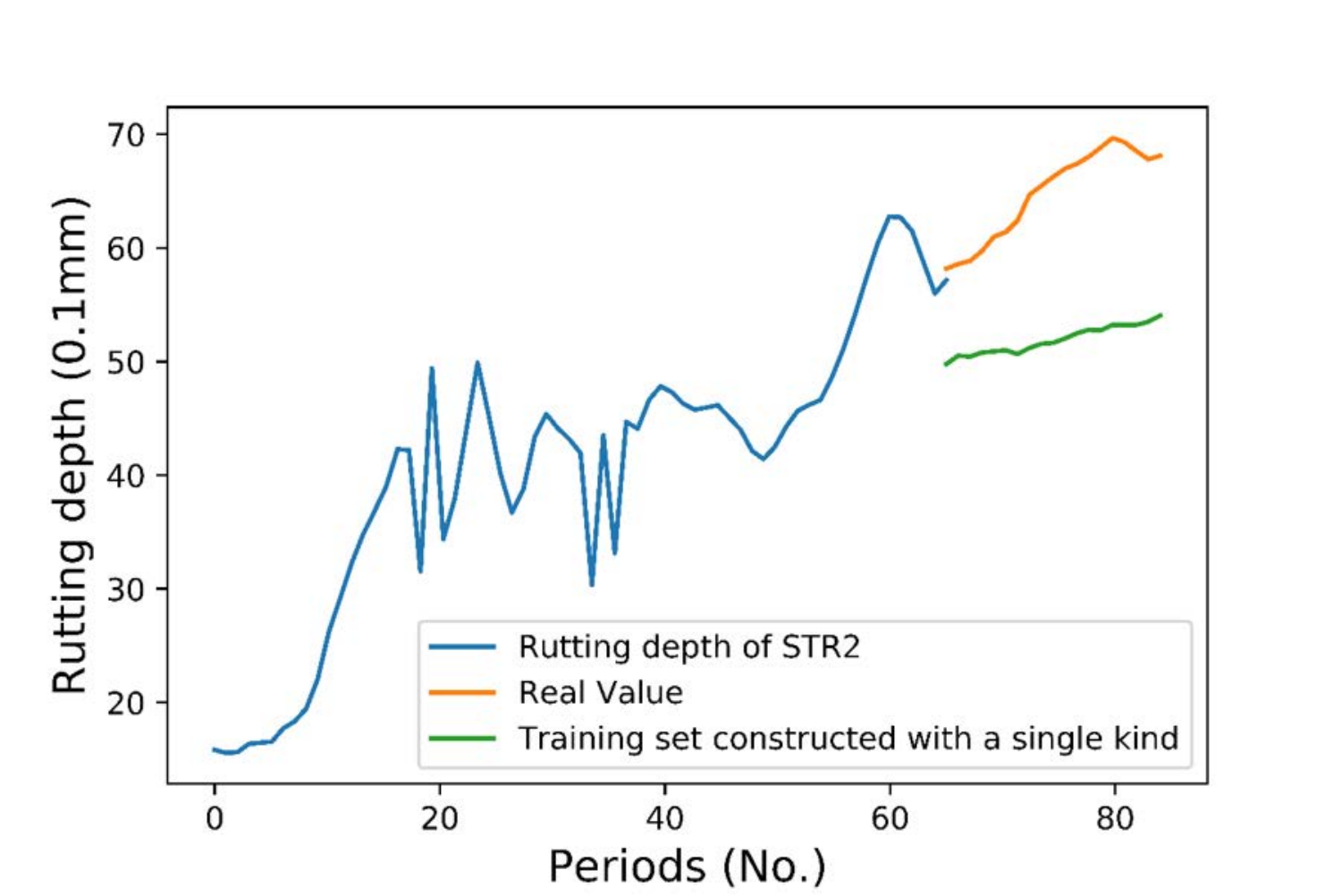}}
    \label{1a}
	  \subfloat[GRU with All Kinds of Asphalt Pavement Data]{
        \includegraphics[width=0.31\linewidth]{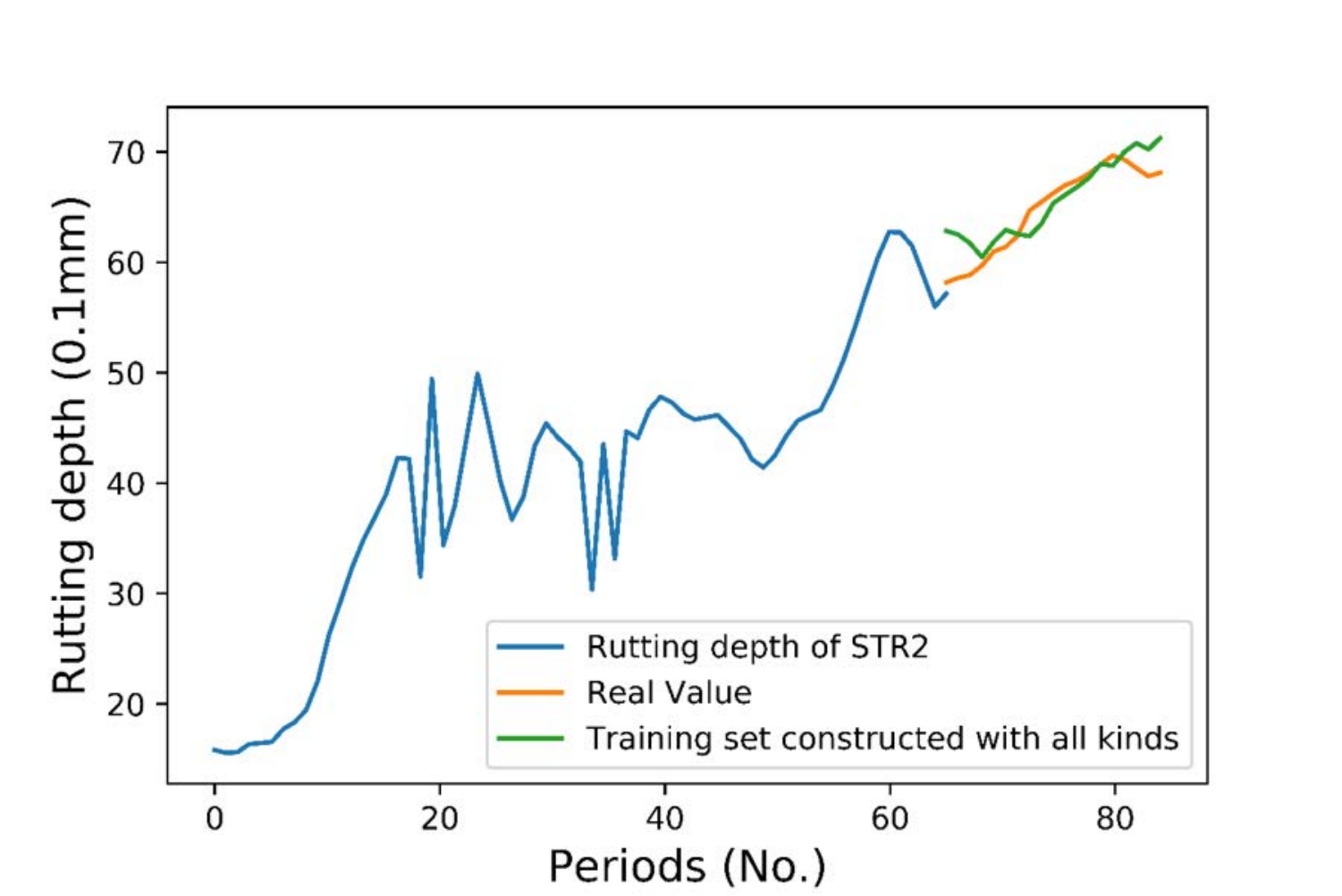}}
    \label{1b}
	  \subfloat[GRU with Equivalent Training Set after Community Detection]{
        \includegraphics[width=0.31\linewidth]{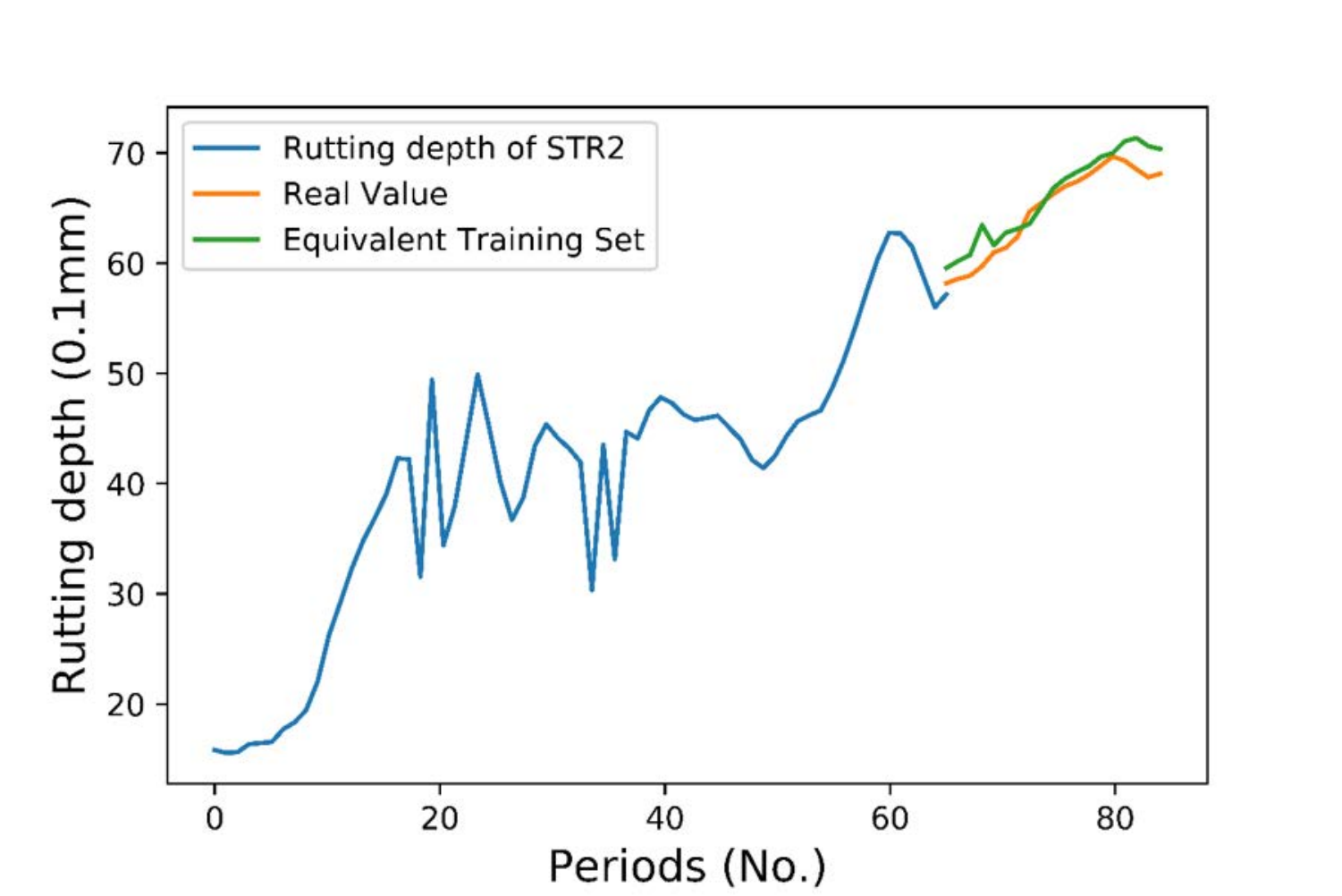}}
    \label{1c}\hfill\\
	  \subfloat[IAPSO-RELM with a Single Kind of Asphalt Pavement Data]{
        \includegraphics[width=0.31\linewidth]{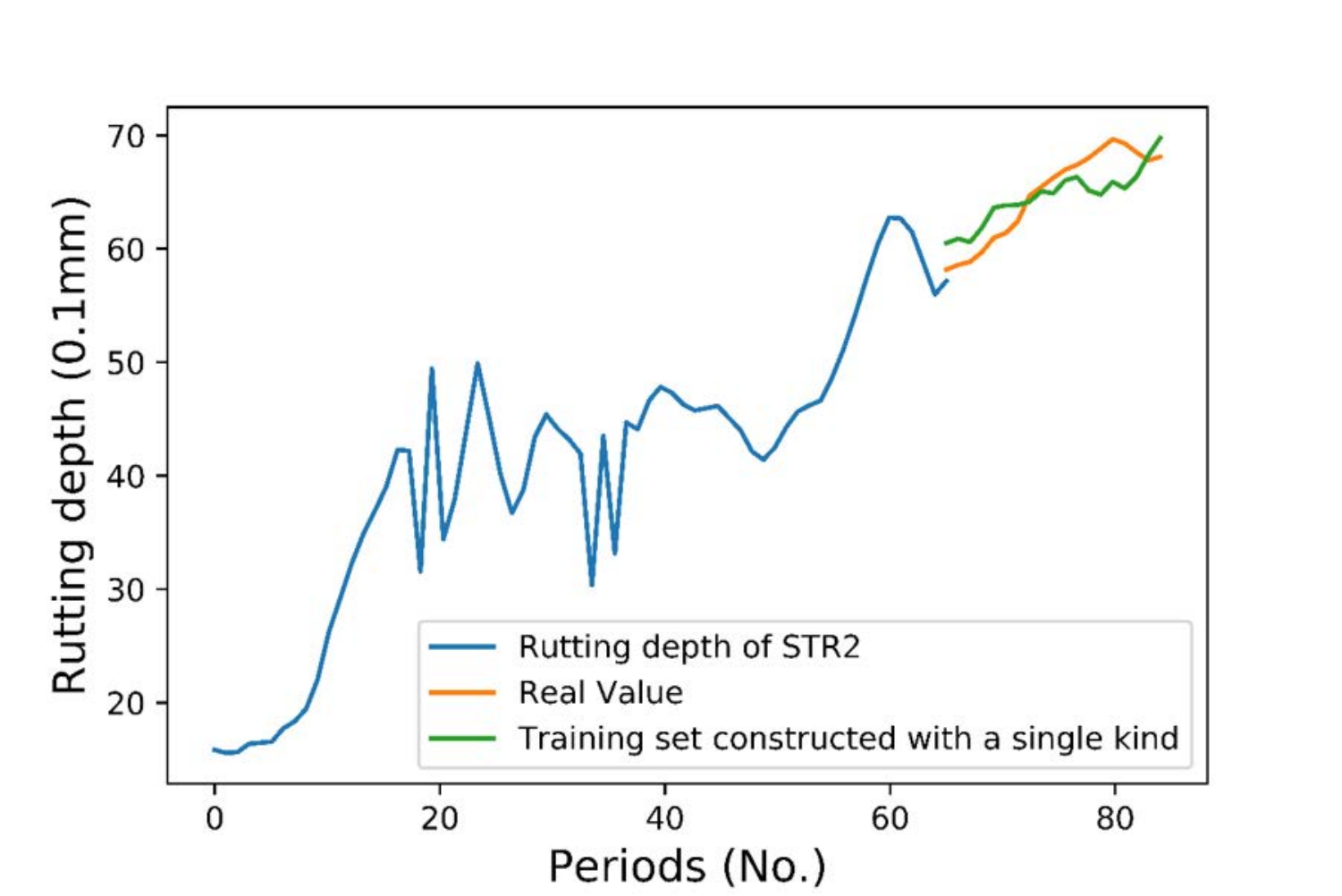}}
     \label{1d} 
	  \label{fig_STR2} 
	  \subfloat[IAPSO-RELM with  All Kinds of Asphalt Pavement Data]{
        \includegraphics[width=0.31\linewidth]{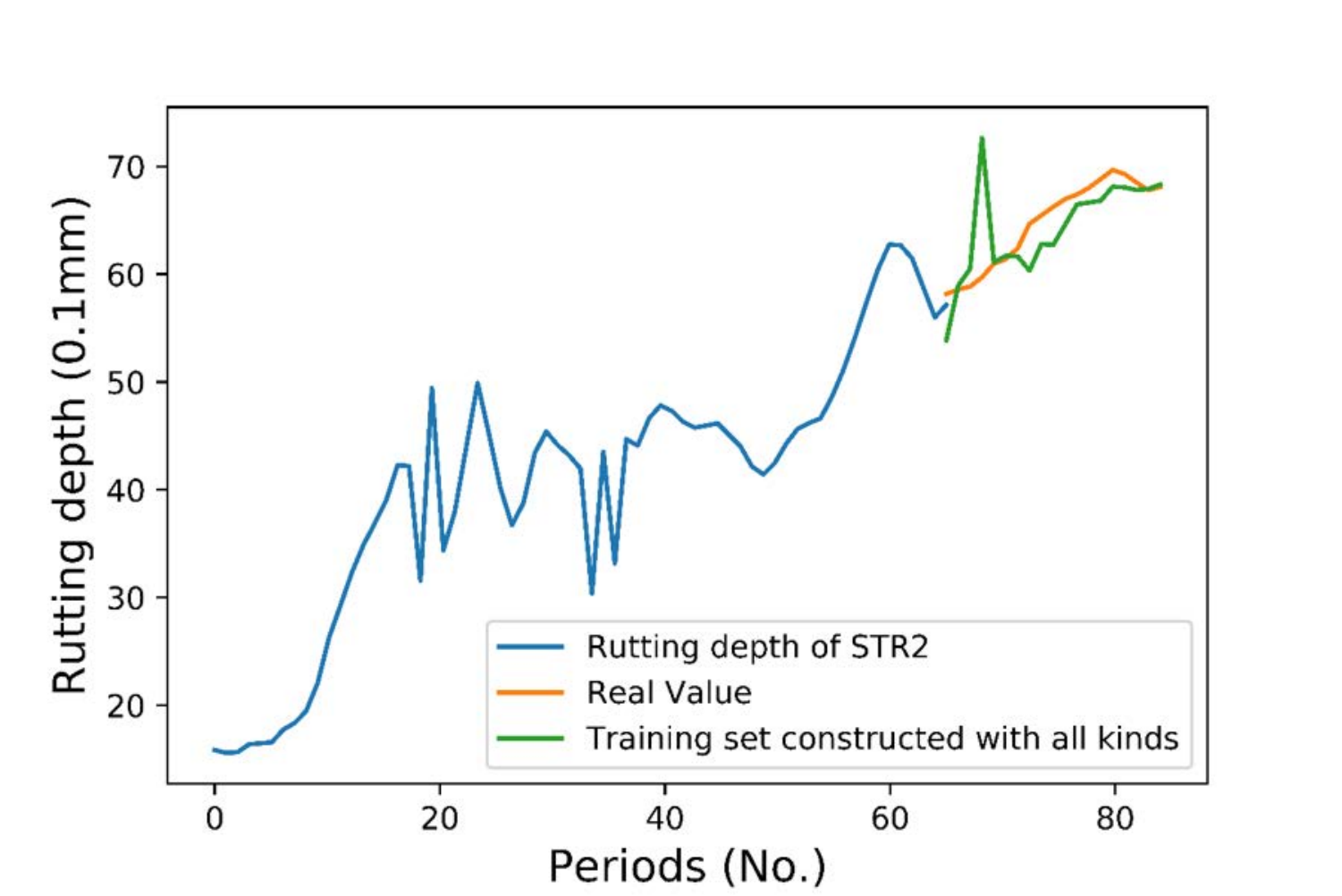}}
    \label{1c}
	  \subfloat[IAPSO-RELM with Equivalent Training Set after Community Detection]{
        \includegraphics[width=0.31\linewidth]{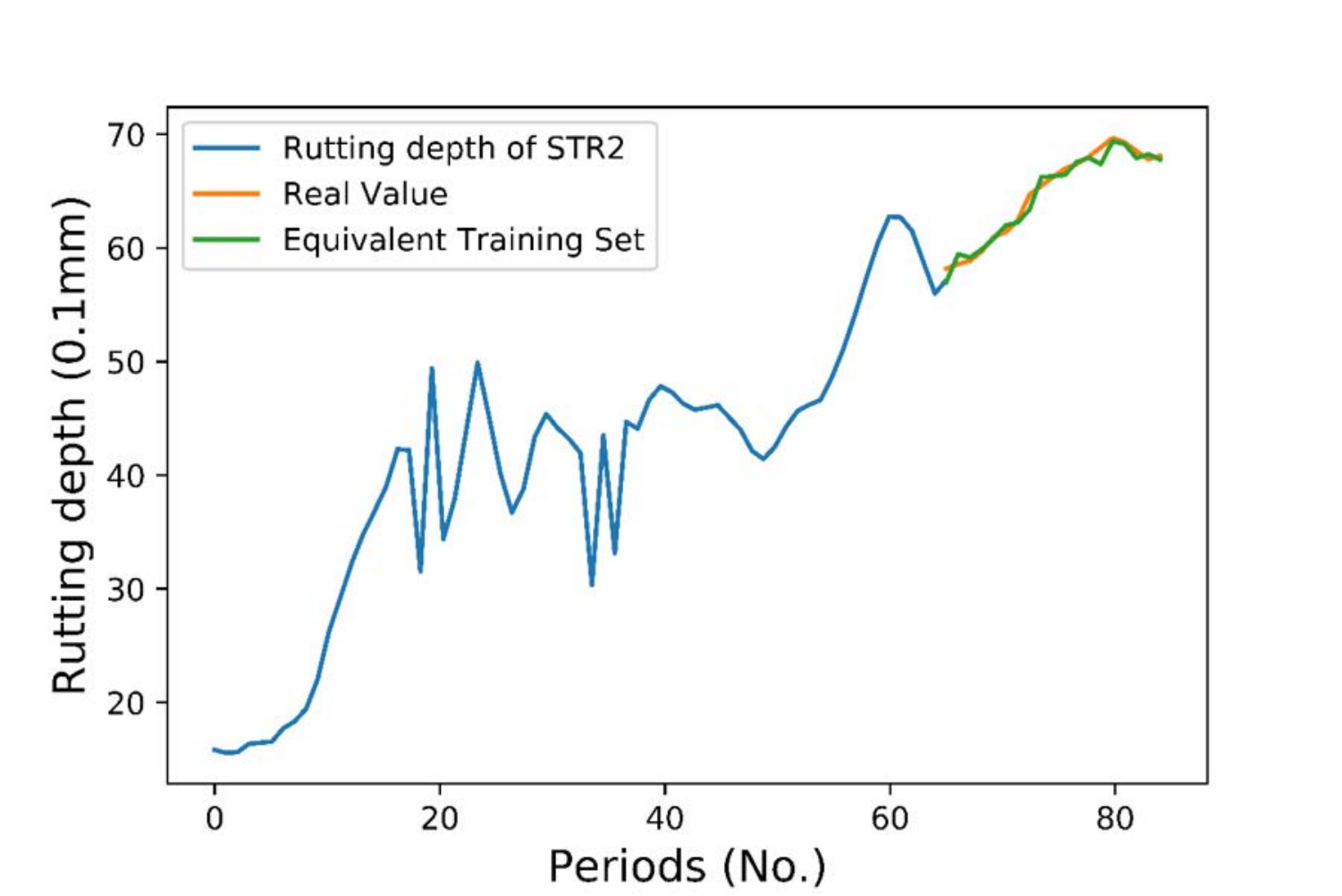}}
     \label{1d} 
	  \caption{Rutting depth prediction results of STR2 with IAPSO-RELM and GRU.}
	  \label{fig_STR2} 
\end{figure*}

\subsubsection{Performance of Louvain algorithm for community detection}
To validate the performance of the complex network approach, we compared it with other clustering methods. Further , we compared our results with density-based noise-space clustering (DBSCAN) \cite{ester1996density}, K-means clustering algorithms \cite{macqueen1967classification} (Number of clusters=3) , Gaussian Mixture Model (GMM) \cite{bishop2006pattern} (Number of clusters=3) , and ordering points to identify the clustering structure (OPTICS) \cite{ankerst1999optics}. We use commonly used metrics to assess the quality of clustering. Due to the data-driven nature of this task, there are no grouped asphalt pavement real-world labels. We focus on intrinsic metrics \cite{wang2009cvap}: silhouette coefficient (SC), Calinski-Harabaz index (CHI) and Davies-Bouldin Index (DBI).The SC is suitable for situations where the actual class information is unknown. For a sample set, its silhouette coefficient is the average of all sample silhouette coefficients. The value range of the silhouette coefficient is $[-1, 1]$. The closer the samples of the same category are and the farther the samples of different categories are, the greater the value of the silhouette coefficient. The DBI, also known as the classification fitness index, calculates the average similarity between a group and the mean of all other groups. The smaller the value, the better the clustering result. CHI, this evaluation index is simple and straightforward to calculate. The smaller the covariance of the data within the category, the larger the covariance between the categories, the better, the higher the CHI. The performance of the clustering algorithm is shown in Table \ref{table_clustering}.

\begin{table}[h]
	\caption{The performance of the clustering algorithm}
	\centering
	\label{table_clustering}
	\begin{tabular}{llll}  	
		\toprule   	
		Algorithm               & SC & DBI & CHI\\
		\midrule 
        Proposed method        & \textbf{0.4471} & 0.7193 &\textbf{37.9102}\\
        DBSCAN        & 0.1744  & 1.0047 &4.6075\\ 
        K-means       & 0.3466  & \textbf{0.5752} &21.6640\\ 
        GMM       & 0.0015  & 2.0996 &5.9596\\ 
        OPTICS       &  0.1889 & 1.6063 &6.4478\\ 
		\bottomrule  
	\end{tabular}
\end{table}

As shown in the Table \ref{table_clustering}, the method proposed in this paper has the best performance on SC and CHI indicators. Compared with the K-means algorithm, only the DBI is slightly higher, but the K-means algorithm needs to pre-set the number of cluster categories. The method proposed in this paper does not need to pre-set the number of clustering categories and can achieve an excellent clustering effect.

\subsubsection{Performance of equivalent training set for the same class after community detection}
To verify the improvement in prediction accuracy of equivalent training set after community detection, the training set is constructed with a single kind of asphalt pavement data, and the training is constructed with all kinds of asphalt pavement data for comparison respectively. In this paper, IAPSO-RELM, RELM, ELM, ELM with RBF Kernel (KELM-RBF) \cite{iosifidis2015kernel}, Long short-term memory (LSTM) \cite{hochreiter1997long}, MLP \cite{ghritlahre2018exergetic}, Gated Recurrent Unit (GRU) \cite{cho2014learning} are used as test algorithms for 19 kinds of asphalt pavements. The performance of algorithms is shown in the Tables \ref{table_Single}-\ref{table_CD}.

\begin{table}[h]
	\caption{The performance comparison on the  training set which is constructed with a single kind of asphalt pavement data}
	\centering
	\label{table_Single}
	\begin{tabular}{lllll}  	
		\toprule   	
		Algorithm  & Average MAPE & Average RMSE & Average MAE \\
		\midrule 
        MLP        & 8.73\%    & 6.592604  & 6.224704 \\
        IAPSO-RELM & \textbf{3.66\%}    & \textbf{3.078302}  & \textbf{2.650691 }\\
        ELM        & 9.84\%    & 7.898636  & 7.501865 \\
        KELM-RBF   & 6.55\%    & 5.25545   & 4.827216 \\
        RELM       & 7.83\%    & 6.207217  & 5.751715 \\
        LSTM       & 23.45\%   & 17.71548  & 17.46174 \\
        GRU        & 15.13\%   & 11.48212  & 11.19007 \\
		\bottomrule  
	\end{tabular}
\end{table}

\begin{table}[h]
	\caption{The performance comparison on the training set which is constructed with all kinds of asphalt pavement data}
	\centering
	\label{table_All}
	\begin{tabular}{lllll}  	
		\toprule   	
		Algorithm  & Average MAPE & Average RMSE & Average MAE \\
		\midrule 
        MLP        & 4.27\% & 3.8522825 & 2.841063  \\
        IAPSO-RELM & \textbf{2.45\%} & \textbf{2.2968817} & \textbf{1.630076}  \\
        ELM        & 5.22\% & 4.3113605 & 3.8856632 \\
        KELM-RBF   & 5.36\% & 4.6498991 & 4.0058338 \\
        RELM       & 3.97\% & 3.2541017 & 2.8225649 \\
        LSTM       & 4.75\% & 3.6816538 & 3.1584351 \\
        GRU        & 4.95\% & 3.9618938 & 3.2667651\\
		\bottomrule  
	\end{tabular}
\end{table}

\begin{table}[h]
	\caption{The performance comparison on the training set which is equivalent training set after community detection}
	\centering
	\label{table_CD}
	\begin{tabular}{lllll}  	
		\toprule   	
		Algorithm  & Average MAPE & Average RMSE & Average MAE \\
		\midrule 
        MLP        & 4.13\%                   & 3.748924                 & 2.7051608                \\
        IAPSO-RELM & \textbf{1.94\% }         &\textbf{1.7420248  }    & \textbf{1.3625498 }              \\
        ELM        & 4.73\%                   & 4.0241488                & 3.6019975               \\
        KELM-RBF   & 4.54\%                   & 3.8532062                & 3.2950194               \\
        RELM       & 3.61\%                   & 2.9645706                & 2.5723887               \\
        LSTM       & 3.53\%                   & 2.7119437                & 2.1632274               \\
        GRU        & 2.71\%                   & 2.5059797                & 1.8047246              \\
		\bottomrule  
	\end{tabular}
\end{table}

As shown in Tables \ref{table_Single}-\ref{table_CD}, it is not difficult to find that for neural network algorithms, among the three different dataset settings, using the equivalent dataset has the highest accuracy. Compared with the single-type asphalt rutting dataset, the Average MAPE trained with the equivalent dataset has the most significant improvement, the minor improvement, and the average improvement of 84.97\%, 30.68\%, and 56.94\%, respectively. Average RMSE maximum improvement, minimum improvement, and average improvement are 84.69\%, 26.68\%, and 53.26\%, respectively. Average MAE maximum improvement, minimum improvement, and average improvement are 87.61\%, 31.74\%, and 58.69\%, respectively. Compared with various asphalt rutting datasets, the Average MAPE trained with the equivalent dataset has the most major improvement; the minimum improvement and the average improvement are 45.15\%, 3.22\%, and 18.38\%, respectively. Average RMSE maximum improvement, minimum improvement, and average improvement are 36.75\%, 2.68\%, and 17.52\%, respectively. Average MAE maximum improvement, minimum improvement, and average improvement are 44.75\%, 4.78\%, and 18.77\%, respectively. Training with equivalent datasets can effectively improve the accuracy of neural network algorithms. The complex asphalt pavement network is effectively clustered by the Louvain algorithm, which enlarges the training set and reduces the noise of the training set.

Taking STR2 as an example, the prediction results of IAPSO-RELM and GRU on different training set construction methods are shown in Fig. \ref{fig_STR2}. As shown in Fig. \ref{fig_STR2}, when using the equivalent training set for rutting depth prediction of asphalt pavement STR2, the proposed algorithm  can achive higher accuracy and better prediction of rutting depth.

\begin{figure}[h]
	\centering
	\includegraphics[scale=0.6]{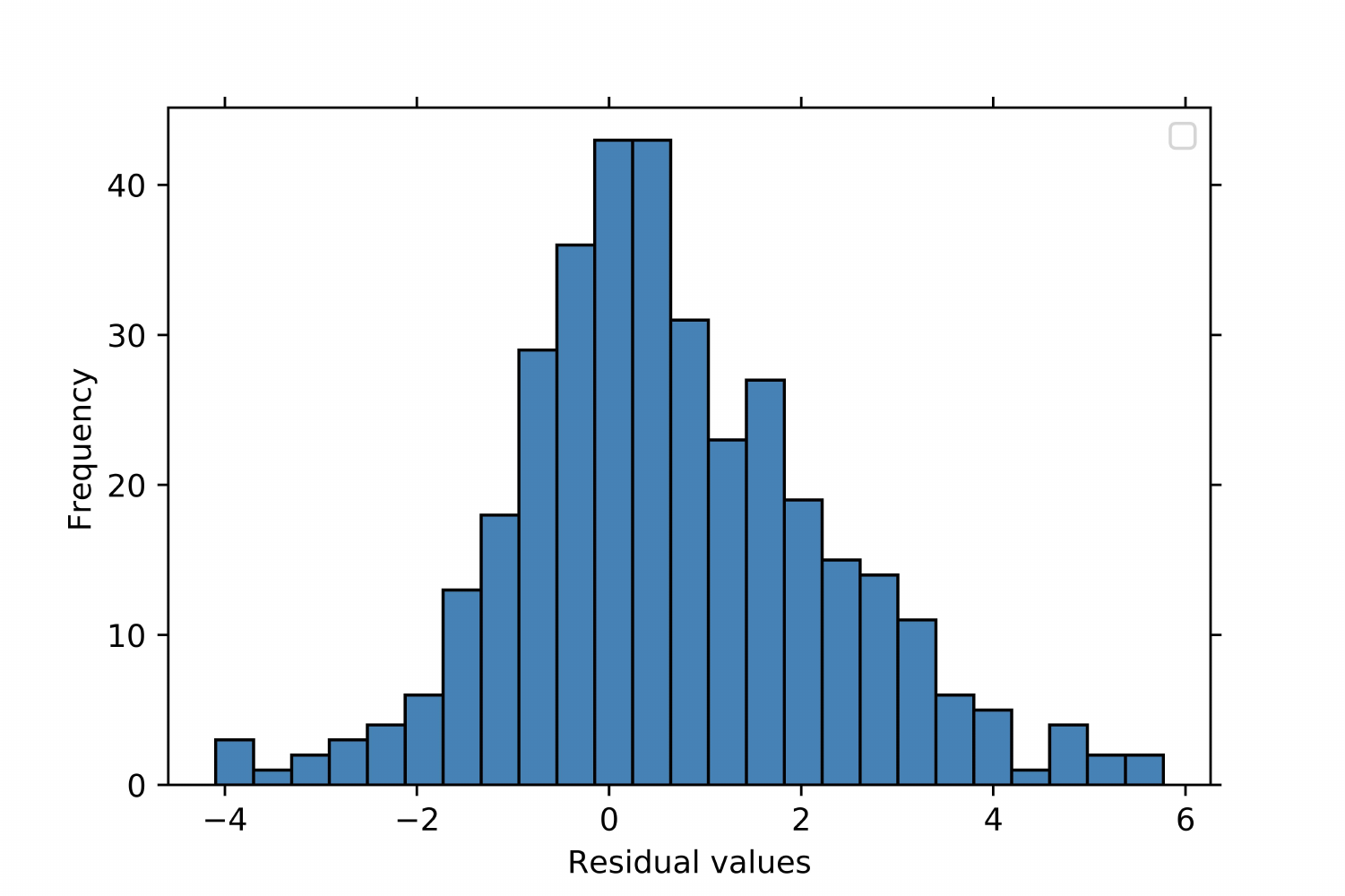}
	\caption{Histogram of the predicted residuals of the model in this paper}
	\label{fig_ZFT}
\end{figure}

\begin{figure}[h]
	\centering
	\includegraphics[scale=0.6]{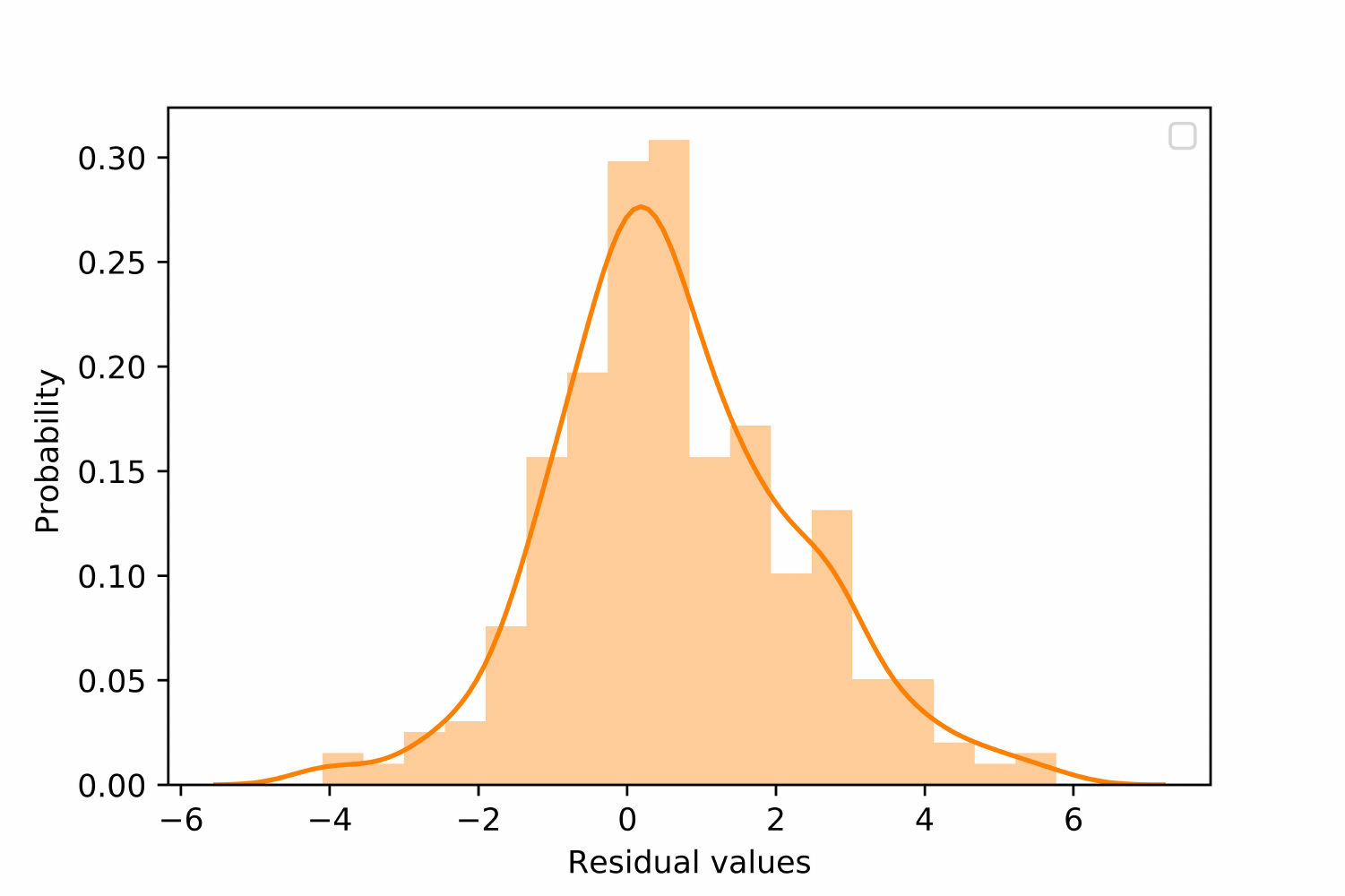}
	\caption{The predicted residual probability distribution of the model in this paper}
	\label{fig_GLMD}
\end{figure}

If the rutting depth short-time prediction model is satisfactory, the distribution of residuals should be expected with zero as the mean. Therefore, the test of residuals includes the test of normal distribution of residuals and the test of normal overall mean distribution. The frequency histogram of the residual distribution of the IAPSO-RELM rutting depth short-time prediction model predicts 19 pavement ruts using an equivalent training set, and the histogram of the residuals of the prediction results are shown in Fig.\ref{fig_ZFT}. The probability density plot of the residuals is shown in Fig.\ref{fig_GLMD}. 

Fig.\ref{fig_GLMD} shows that the prediction residuals of the model in this paper are close to the normal distribution. In order to better test the distribution of the model prediction residuals, the K-S test \cite{young1977proof} was used to test whether the model's residuals in this paper were average to verify the model's reliability. Given the significance level $a=0.05$, the statistic calculated by the K-S test is 0.06775, and the p-value is 0.0694 ($>0.05$) does not reject the original hypothesis, which means that the predicted residuals of this model obey a normal distribution.

\subsection{Performance of IAPSO-RELM in Uncertainty}
Discussing the uncertainty of machine learning models can help to understand the model's performance and help to create believable artificial intelligence. The uncertainty in the short-time prediction of rut depth for the model proposed in this paper mainly stems from the random initialization of the ELM algorithm for the weights, which leads to uncertainty in the prediction. For model uncertainty, predictions are made by conducting multiple experiments with different model initializations. The variance of multiple prediction results is calculated to represent the uncertainty of the model, and the calculation method is as follows:
\begin{equation}
    \label{delta}
    \overline{\delta^2} = \frac{1}{M\times T}\sum^M_i\sum^T_j (x_i^j-\mu^i)^2
\end{equation}
where $T$ is the number of trials, $M$ is the number of predicted samples, $x_i^j$ is the predicted value of the $i$-th sample in the $j$-th trial, and $\mu^i$ is the average value of the predicted value of the $i$-th sample.

To verify the improvement of the certainty of the IAPSO-RELM model, the MLP, ELM, RELM, and IAPSO-RELM models are trained and predicted 100 times with different initial weights, respectively. The uncertainty of the models on the train and test sets was calculated by Eq.\ref{delta}. We will predict the rutting depth of  asphalt pavements in Group I. The lower the value, the lower the uncertainty of the model. The detailed performance is shown in the Table \ref{table_Uncertainty}.

\begin{table}[h]
	\caption{The performance comparison on uncertainty of different Algorithms}
	\centering
	\label{table_Uncertainty}
	\begin{tabular}{lll}  	
		\toprule   	
		Algorithm           & $\overline{\delta^2}$ of test set   & $\overline{\delta^2}$ of train set \\ 
		\midrule   
		MLP                 & 25.9547                       & 6.7201 \\
        ELM                 & 25.1153                       & 4.3990 \\
        RELM                & 19.3632                       & 2.4814  \\
        IAPSO-RELM          & \textbf{3.7318}               & \textbf{0.2592}  \\
		\bottomrule  
	\end{tabular}
\end{table}

Table \ref{table_Uncertainty} shows that the weights of the ELM algorithm are randomly generated, resulting in relatively high uncertainty on both the training and test sets. The ELM algorithm with residual correction shows a significant improvement in uncertainty. Using IAPSO to optimize the RELM algorithm, the uncertainty of the model is greatly improved, and the prediction is more stable. In the follow-up work, the uncertainty indicator will be used as an optimization indicator to improve the algorithm.

\subsection{Performance of IAPSO}
\begin{figure}[h]
	\centering
	\includegraphics[scale=0.55]{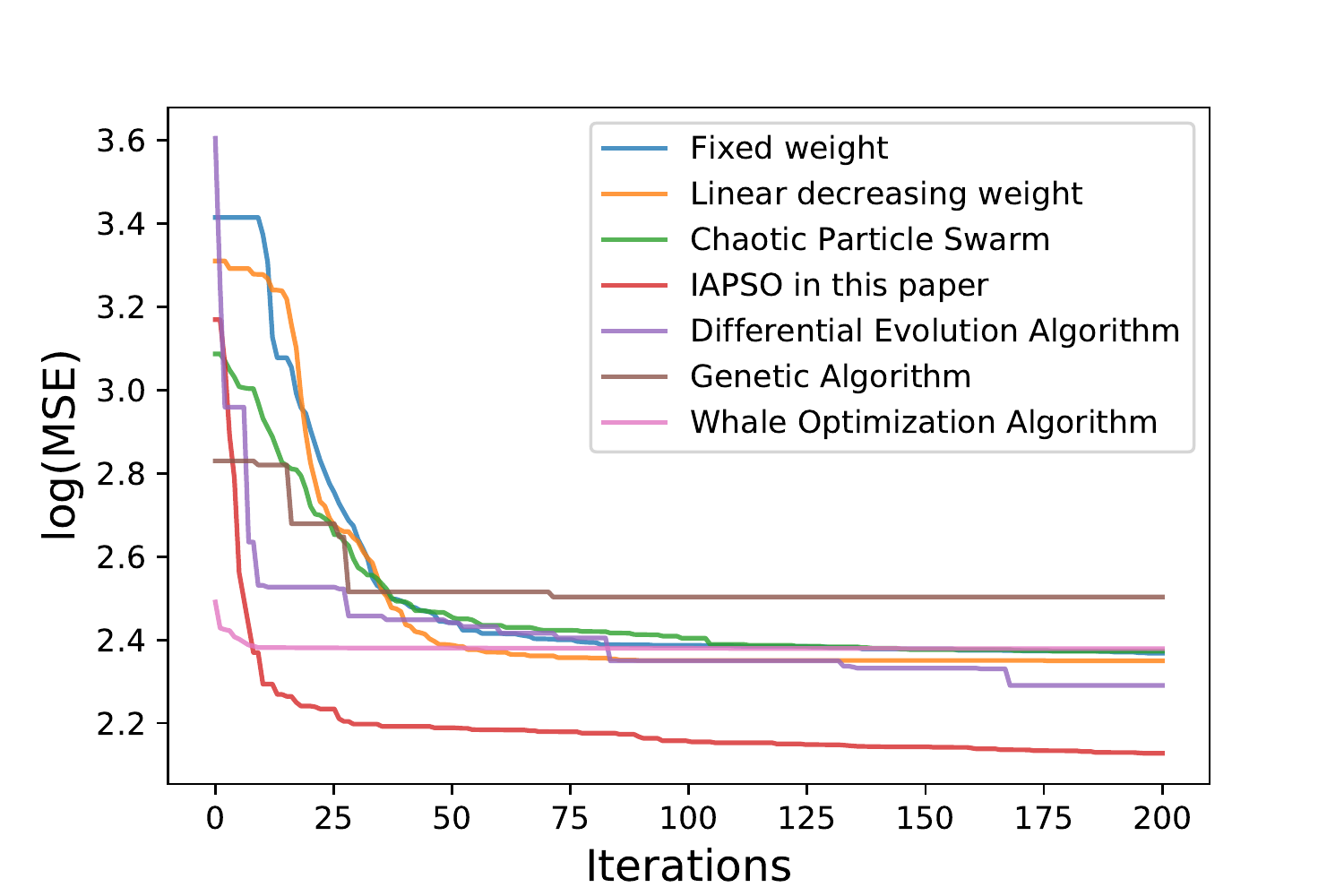}
	\caption{Fitness graph of different evolutionary algorithms. }
	\label{fig_CPSO_C}
	\label{fig:visual_smap_gt}
\end{figure}

We will predict the rutting depth of  asphalt pavements in Group I.  Firstly, the performance of IAPSO is analyzed. For IAPSO optimization, the initial parameters are set as follows: maximum iteration $T=100$, number of population $M=30$, $ x_{max} = 1 $, $ v_{max} = 1 $ , $w_{\max }=0.9$ , $w_{\min }=0.4$, $c_{1}^{\max}=2$, $c_{2}^{\max}=2$, $c_{1}^{\min }=0.8$ and $c_{2}^{\min }=0.8$. Moreover, the 
IAPSO is compared with some existing PSO algorithms with fixed weight \cite{kennedy1995particle}, linear decreasing weight \cite{shi1999empirical} and chaotic optimization \cite{Li2022CPSO} in the same initial particle distribution. Classical evolutionary algorithms such as Genetic Algorithm \cite{whitley1994genetic}, Differential Evolution Algorithm \cite{wang2012annual}, and Whale Optimization Algorithm \cite{mirjalili2016whale} are also used as comparison algorithms. The corresponding fitness graph of different algorithms is shown in Fig.\ref{fig_CPSO_C}.  In the iterative process of these algorithms, we make some of the particles in the swarm mutate and escape from the initial range, to avoid falling into the local optimum.  The performance comparison is shown in Table \ref{table_CPSO_C}.

\begin{table}[h]
	\caption{Optimization comparison among different evolutionary algorithms}
	\centering
	\label{table_CPSO_C}
	\begin{tabular}{ll}  	
		\toprule   	
		Algorithm                       & MSE \\ 
		\midrule   
		Fixed weight                        & 10.678230310134028\\
        Linear decreasing weight            & 10.482436613266115\\
        Chaotic optimization                & 10.728301256131255 \\
        IAPSO in this paper                 & \textbf{8.395501284204803} \\
        Differential Evolution Algorithm    & 9.880636295563475 \\
        Genetic Algorithm                   & 12.219300233403883 \\
        Whale Optimization Algorithm                   & 10.795870458149949\\
		\bottomrule  
	\end{tabular}
\end{table}

The results show that the IAPSO algorithm proposed in this paper has the smallest fitness function value for all particles in the last iteration compared with other PSO algorithms. It uses an independent adaptive parameter update method in the search to jump out of the local optimum and obtain a lower fitness function value. IAPSO demonstrates the effectiveness of the IAPSO algorithm by improving the training accuracy by 20\% over other PSO algorithms.

\subsection{Rutting Depth Prediction for Different Types of Asphalt Pavements}
To illustrate the generality of our algorithm, we considered the prediction of rutting depth under load tests on 19 asphalt pavements. To validate the effectiveness of the IAPSO-RELM algorithm with the equivalent training set after community detection in the short-time prediction task of rutting depth, we compare it with commonly used models for predicting rutting depth, such as XGBoost \cite{chen2016xgboost}, SVR \cite{smola2004tutorial}, K-Nearest Neighbor (KNN) \cite{yang2019evaluation}, Random Forest (RF) \cite{smith2013comparison}, Decision Tree (DT) \cite{kurt2018classification}, RELM, ELM, KELM-RBF, LSTM, MLP, and GRU as baseline models for comparison. The baseline models using machine learning algorithms are implemented using Scikit learn \cite{pedregosa2011scikit} library for python, with the default, recommended parameters. For LSTM and GRU, we set up nine input nodes, 128 hidden nodes, and one output node. The model is then trained with the same training dataset with a learning rate of 0.01. We solved the model parameters using a batch size of 128 and a maximum epoch of 500. The stochastic gradient descent (SGD) algorithm is used for solving the parameters of models. All comparison algorithms are also tested using the same equivalent training set.

\begin{figure}[H]
    
    \centering
	    \subfloat[RMSE of Comparison Algorithms ]{
            \includegraphics[width=0.85\linewidth]{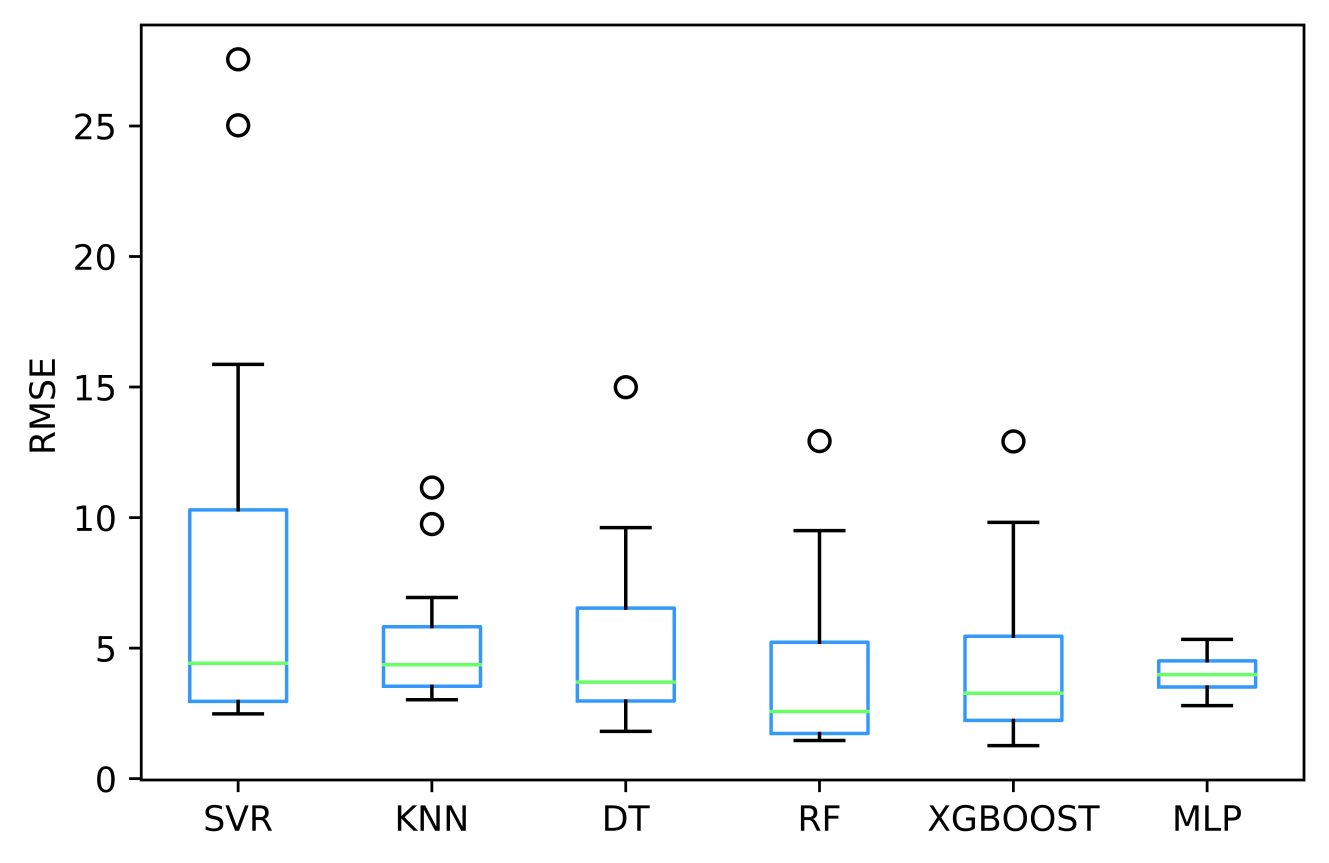}}\\
	    \subfloat[RMSE of Comparison Algorithms ]{
            \includegraphics[width=0.85\linewidth]{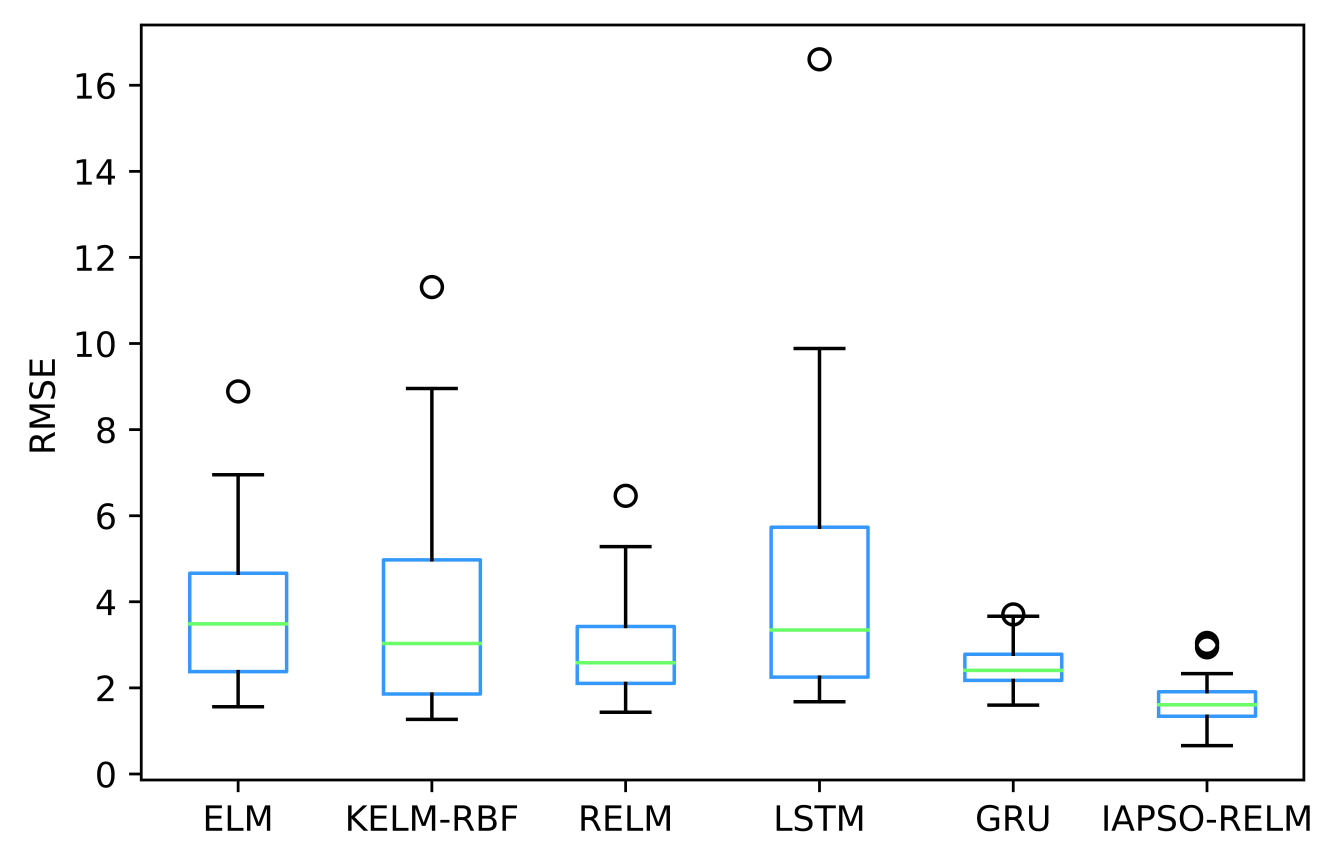}}
    \caption{Boxplots of RMSE with different algorithms}
    \label{fig_BoxRMSE}
\end{figure}

\begin{figure}[H]
    \centering
    
	  \subfloat[MAPE of Comparison Algorithms ]{
        \includegraphics[width=0.85\linewidth]{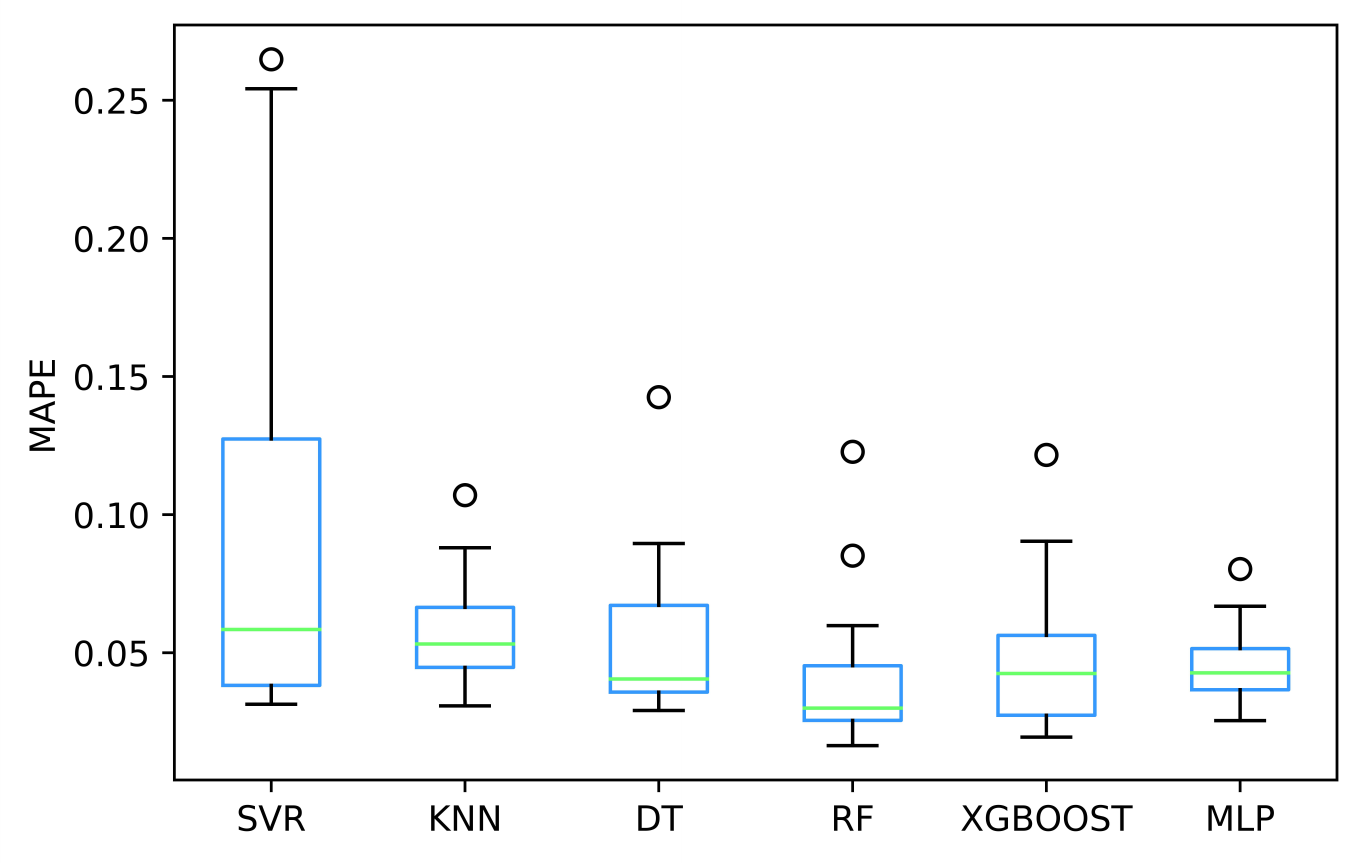}}\\
	  \subfloat[MAPE of Comparison Algorithms ]{
        \includegraphics[width=0.85\linewidth]{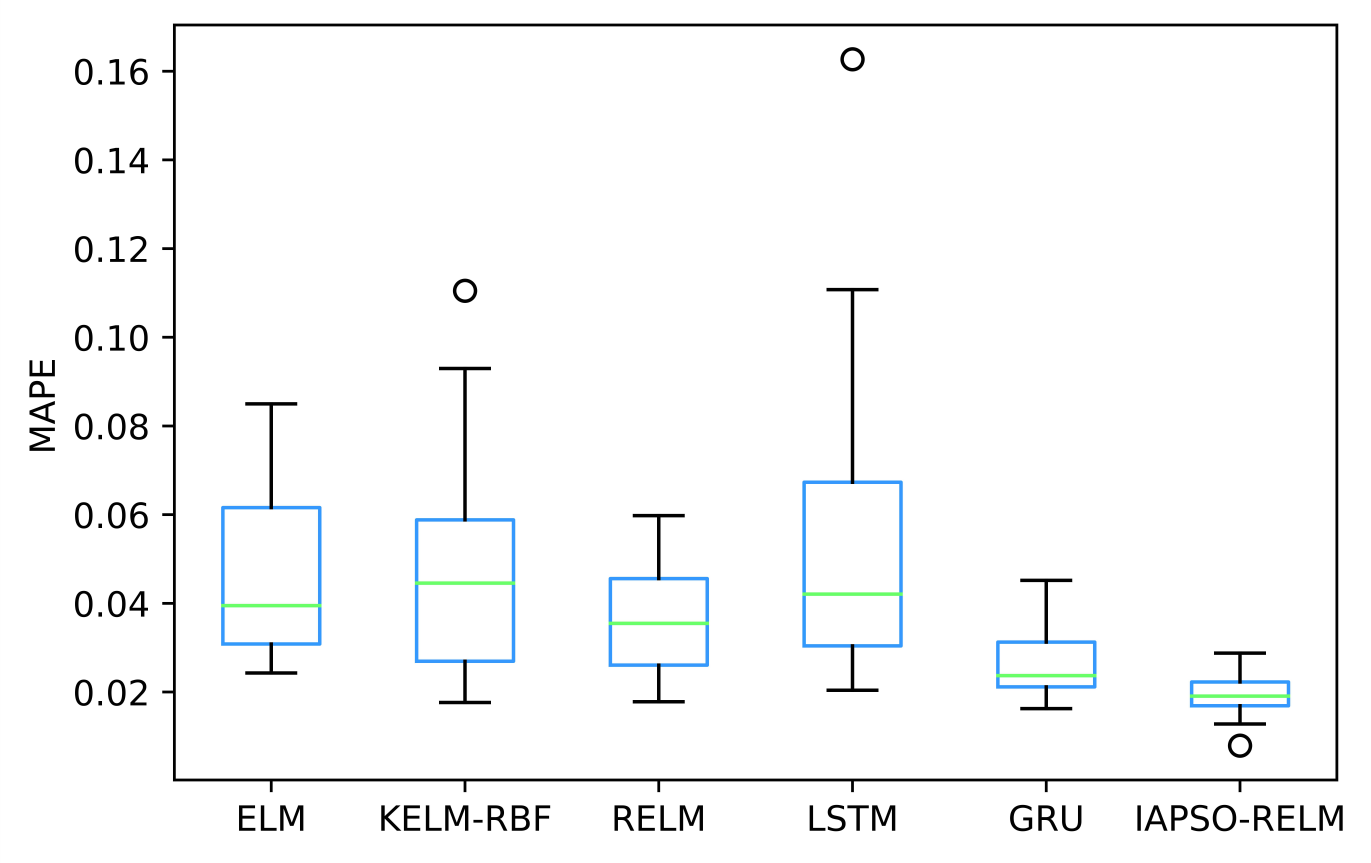}}
     \caption{Boxplots of MAPE with different algorithms}
     \label{fig_BoxMAPE}
\end{figure}

In the paper, the performances of the IAPSO-RELM algorithm with the equivalent training set after community detection, XGBoost, SVR, KNN, RF, RELM, ELM, KELM-RBF, LSTM, MLP, and GRU on RMSE, MAE, and MAPE are compared as shown in Fig.\ref{fig_BoxRMSE}-\ref{fig_BoxMAPE}.

\begin{figure}[H]
    \centering
    \label{fig_BoxMAE}
	  \subfloat[MAE of Comparison Algorithms ]{
        \includegraphics[width=0.85\linewidth]{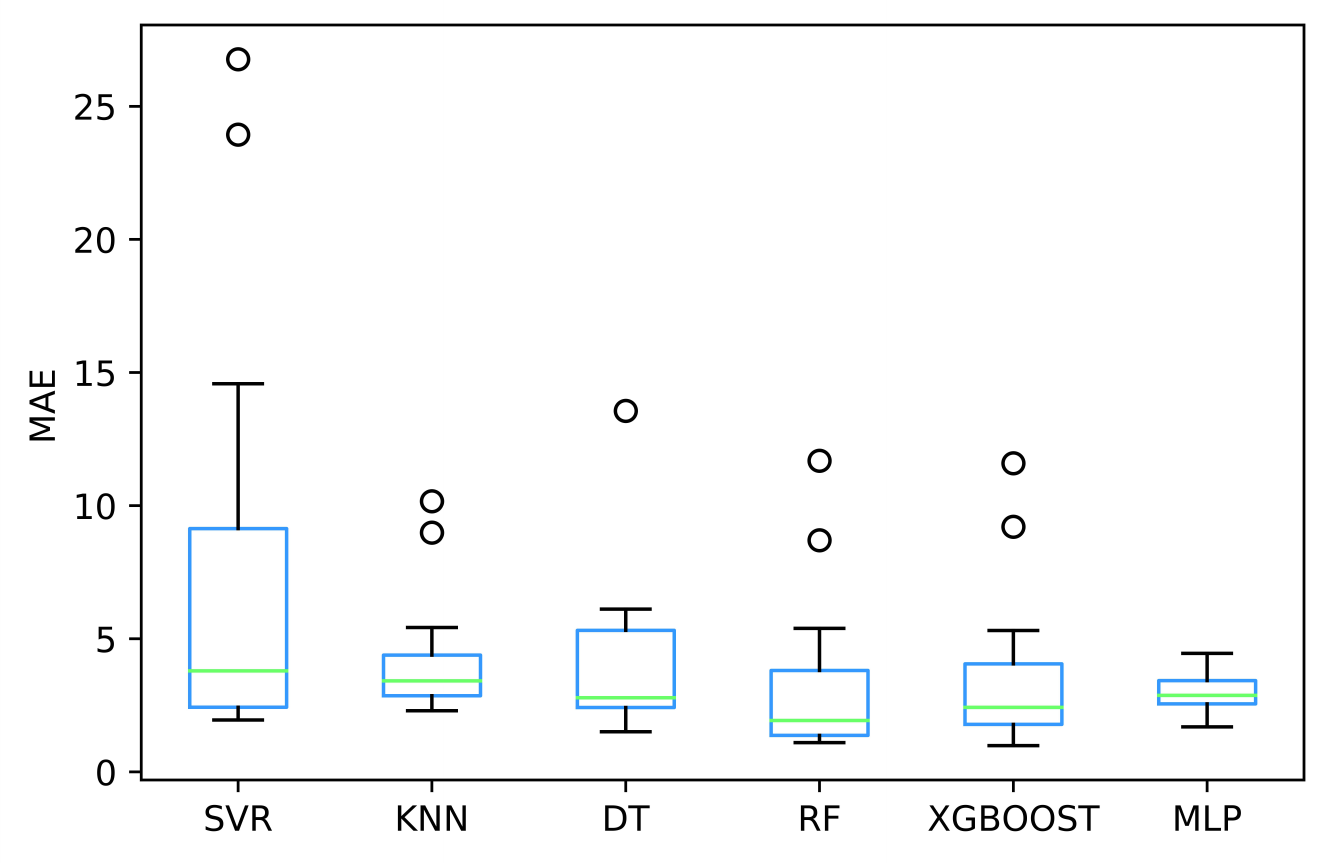}}\\
	  \subfloat[MAE of Comparison Algorithms ]{
        \includegraphics[width=0.85\linewidth]{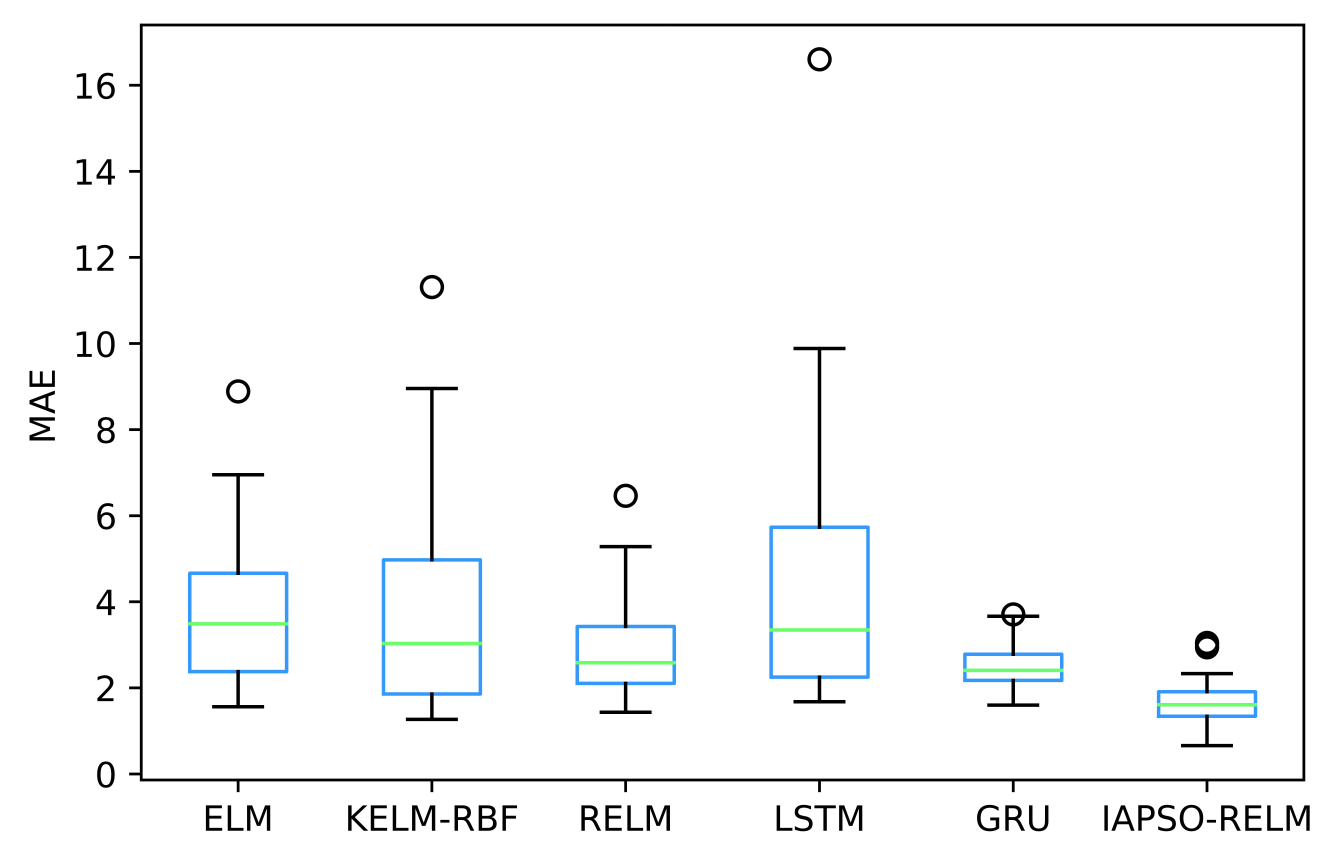}}
     \caption{Boxplots of MAE with different algorithms}
     \label{fig_BoxMAE}
\end{figure}

\begin{table}[h]
	\caption{The performance of algorithms in RMSE, MAPE and MAE}
	\centering
	\scriptsize
	\label{table_IAPSO}
	\begin{tabular}{cccc}  	
		\toprule   	
		Algorithm                & Average MAPE   & Average RMSE  & Average MAE   \\
		\midrule   
        SVM                      & 9.28\% & 8.027 & 7.188 \\
        KNN                      & 5.75\% & 5.075 & 4.124 \\
        DT                       & 5.51\% & 5.036 & 4.053 \\
        RF                       & 4.01\% & 3.860 & 3.111 \\
        XGBOOST                  & 4.60\% & 4.324 & 3.473 \\
        MLP                      & 4.53\% & 4.049 & 3.005 \\
        ELM                      & 4.73\% & 4.024 & 3.602 \\
        KELM-RBF                 & 4.54\% & 3.853 & 3.295 \\
        RELM                     & 3.61\% & 2.965 & 2.572 \\
        LSTM                     & 3.53\% & 2.712 & 2.163 \\
        GRU                      & 2.71\% & 2.506 & 1.805 \\
        IAPSO-RELM in this paper & \textbf{1.94\%} & \textbf{1.742} & \textbf{1.363}\\
        \bottomrule  
	\end{tabular}
\end{table}

The comparative results of the algorithms are shown in Table \ref{table_IAPSO}, which indicates that the algorithm proposed has the best performance among all the conducted algorithms in RMSE, MAE, and MAPE for the rutting depth short time prediction task on 19 asphalt pavement datasets. RELM algorithm outperforms the XGBOOST, KNN, SVR, ELM, KELM-RBF, MLP, and RF algorithms in terms of prediction accuracy. The IAPSO algorithm is used to optimize the parameters of the RELM algorithm, and the accuracy is further improved. Compared with traditional machine learning algorithms and deep learning algorithms such as LSTM and GRU, IAPSO-RELM algorithm has improved the RMSE by a maximum of 78.30\%, a minimum of 30.49\% and an average of 54.54\%; it has improved the MAE by a maximum of 81.05\%, a minimum of 24.50\% and an average of 55.94\%; it has improved the MAPE by a maximum of 79.05\%, a minimum of 28.38\% and 55.62\% on average.

\section{Conclusion}
\label{sec:Conclusion}
In response to the low rutting data of single asphalt pavement in the RIOHTrack test, an algorithm based on complex network community detection was proposed to construct an equivalent data set and improve the prediction accuracy of neural network-like algorithms. Concerning the weak generalization ability and low prediction accuracy of ELM-based prediction models in asphalt pavement rutting prediction, an ELM algorithm using residual correction, RELM, was proposed, and the parameters of the RELM algorithm were optimized using the IAPSO algorithm, named the IAPSO-RELM algorithm.

In this study, the predictive performance of the IAPSO-RELM model was tested using an equivalent dataset using 19 asphalt pavement materials. These data and works were collected and supported by the National Key Research and Development Program of China (2020YFA0714300). The algorithm outperformed other comparative algorithms in terms of RMSE, MAE, and MAPE for all asphalt pavement materials. As the training sample size has a significant impact on the prediction performance of computational intelligence methods, their prediction performance can be improved by increasing the number of training data. The test results show that the IAPSO-RELM model using equivalent data is effective in predicting the rutting depth of asphalt pavements.

By introducing prior knowledge and establishing a strong artificial intelligence model, the accuracy of short-term prediction of asphalt pavement rutting can be improved. To construct a rutting depth complex network of asphalt pavements, different levels of abstraction analyses are carried out for asphalt pavement materials that cannot be measured by themselves. Through community detection and analysis of rutting depth variation of 19 kinds of asphalt pavements, the similarity of different pavement structures was excavated.

In future work, we will focus on considering the combination of metaheuristic optimization and machine learning algorithms for oil and gas systems. Treated like a rutting depth prediction, a credible prediction of key oil and gas system parameters takes into account the simulated environment where multiple environmental parameters are coupled with internal system changes. It provides the basis for the subsequent full-environment simulation and intelligent scheduling of oil and gas systems.


%





\ifCLASSOPTIONcaptionsoff
  \newpage
\fi



%
\bibliographystyle{IEEEtran}
\bibliography{bare_jrnl}

\begin{thebibliography}{10}
\providecommand{\url}[1]{#1}
\csname url@samestyle\endcsname
\providecommand{\newblock}{\relax}
\providecommand{\bibinfo}[2]{#2}
\providecommand{\BIBentrySTDinterwordspacing}{\spaceskip=0pt\relax}
\providecommand{\BIBentryALTinterwordstretchfactor}{4}
\providecommand{\BIBentryALTinterwordspacing}{\spaceskip=\fontdimen2\font plus
\BIBentryALTinterwordstretchfactor\fontdimen3\font minus
  \fontdimen4\font\relax}
\providecommand{\BIBforeignlanguage}[2]{{%
\expandafter\ifx\csname l@#1\endcsname\relax
\typeout{** WARNING: IEEEtran.bst: No hyphenation pattern has been}%
\typeout{** loaded for the language `#1'. Using the pattern for}%
\typeout{** the default language instead.}%
\else
\language=\csname l@#1\endcsname
\fi
#2}}
\providecommand{\BIBdecl}{\relax}
\BIBdecl

\bibitem{gul2019fuzzy}
M.~Gul, A.~Guneri, and S.~M. Nasirli, ``A fuzzy-based model for risk assessment
  of routes in oil transportation,'' \emph{Int. J. Environ. Sci. Technol.},
  vol.~16, no.~8, pp. 4671--4686, 2019.

\bibitem{mir2020impact}
H.~Mir, T.~Abdul Hussain~Ratlamwala, G.~Hussain, M.~Alkahtani, and M.~H. Abidi,
  ``Impact of sloshing on fossil fuel loss during transport,'' \emph{Energies},
  vol.~13, no.~10, p. 2625, 2020.

\bibitem{zheng2020technical}
J.~Zheng, S.~L{\"u}, and C.~Liu, ``Technical system, key scientific problems
  and technical frontier of long-life pavement,'' \emph{Chin. Sci. Bull.},
  vol.~65, no.~30, pp. 3219--3229, 2020.

\bibitem{majidifard2021deep}
H.~Majidifard, B.~Jahangiri, P.~Rath, A.~H. Alavi, and W.~G. Buttlar, ``A deep
  learning approach to predict hamburg rutting curve,'' \emph{Road Mater.
  Pavement Des.}, vol.~22, no.~9, pp. 2159--2180, 2021.

\bibitem{rahman2019effect}
M.~M. Rahman and S.~L. Gassman, ``Effect of resilient modulus of undisturbed
  subgrade soils on pavement rutting,'' \emph{Int. J. Geotech. Eng.}, vol.~13,
  no.~2, pp. 152--161, 2019.

\bibitem{behiry2012fatigue}
A.~E. A. E.~M. Behiry, ``Fatigue and rutting lives in flexible pavement,''
  \emph{Ain. Shams. Eng. J.}, vol.~3, no.~4, pp. 367--374, 2012.

\bibitem{tayfur2007investigation}
S.~Tayfur, H.~Ozen, and A.~Aksoy, ``Investigation of rutting performance of
  asphalt mixtures containing polymer modifiers,'' \emph{Constr. Build.
  Mater.}, vol.~21, no.~2, pp. 328--337, 2007.

\bibitem{nagabhushana2013rutting}
M.~Nagabhushana, D.~Tiwari, P.~Jain \emph{et~al.}, ``Rutting in flexible
  pavement: an approach of evaluation with accelerated pavement testing
  facility,'' \emph{Procedia. Soc. Behav. Sci.}, vol. 104, pp. 149--157, 2013.

\bibitem{polacco2015review}
G.~Polacco, S.~Filippi, F.~Merusi, and G.~Stastna, ``A review of the
  fundamentals of polymer-modified asphalts: Asphalt/polymer interactions and
  principles of compatibility,'' \emph{Adv. Colloid Interface Sci.}, vol. 224,
  pp. 72--112, 2015.

\bibitem{porto2019bitumen}
M.~Porto, P.~Caputo, V.~Loise, S.~Eskandarsefat, B.~Teltayev, and
  C.~Oliviero~Rossi, ``Bitumen and bitumen modification: A review on latest
  advances,'' \emph{Appl. Sci.}, vol.~9, no.~4, p. 742, 2019.

\bibitem{wang2017design}
X.~Wang, ``Design of pavement structure and material for full-scale test
  track,'' \emph{J. Highw. \& Transp. Res. \& Dev.}, vol.~34, no.~6, pp.
  30--37, 2017.

\bibitem{dong2021data}
Q.~Dong, X.~Chen, S.~Dong, and F.~Ni, ``Data analysis in pavement engineering:
  An overview,'' \emph{IEEE Transactions on Intelligent Transportation
  Systems}, 2021.

\bibitem{wang2020key}
X.-d. Wang, G.-l. Zhou, H.-y. Liu, and X.~Qing, ``Key points of riohtrack
  testing road design and construction,'' \emph{J. Highway Transportation Res.
  Dev.}, vol.~14, no.~4, pp. 1--16, 2020.

\bibitem{huang2020surface}
W.~Huang, S.~Liang, and Y.~Wei, ``Surface deflection-based reliability analysis
  of asphalt pavement design,'' \emph{Sci. China-Technol. Sci.}, p.
  1824–1836, 2020.

\bibitem{Liu2022evaluation}
H.~Liu, J.~Cao, W.~Huang, X.~Shi, and X.~Wang, ``Complex network approach for
  the evaluation of asphalt pavement design and construction: a longitudinal
  study,'' \emph{Sci. China Inf. Sci.}, vol.~65, no.~7, p. 172204, 2022.

\bibitem{liu2021rutting}
G.~Liu, L.~Chen, Z.~Qian, Y.~Zhang, and H.~Ren, ``Rutting prediction models for
  asphalt pavements with different base types based on riohtrack full-scale
  track,'' \emph{Constr. Build. Mater.}, vol. 305, p. 124793, 2021.

\bibitem{zhang2021investigation}
W.~Zhang, X.~Chen, S.~Shen, L.~N. Mohammad, B.~Cui, S.~Wu, and A.~Raza~Khan,
  ``Investigation of field rut depth of asphalt pavements using hamburg wheel
  tracking test,'' \emph{J. Transp. Eng. Pt. B-Pavements}, vol. 147, no.~1, p.
  04020091, 2021.

\bibitem{uwanuakwa2020artificial}
I.~D. Uwanuakwa, S.~I.~A. Ali, M.~R.~M. Hasan, P.~Akpinar, A.~Sani, and K.~A.
  Shariff, ``Artificial intelligence prediction of rutting and fatigue
  parameters in modified asphalt binders,'' \emph{Appl. Sci.}, vol.~10, no.~21,
  p. 7764, 2020.

\bibitem{fang2020prediction}
M.~Fang, C.~Han, Y.~Xiao, Z.~Han, S.~Wu, and M.~Cheng, ``Prediction modelling
  of rutting depth index for asphalt pavement using de-noising method,''
  \emph{Int. J. Pavement Eng.}, vol.~21, no.~7, pp. 895--907, 2020.

\bibitem{dettenborn2020pavement}
T.~Dettenborn, A.~Hartikainen, and L.~Korkiala-Tanttu, ``Pavement maintenance
  threshold detection and network-level rutting prediction model based on
  finnish road data,'' \emph{J. Infrastruct. Syst.}, vol.~26, no.~2, p.
  04020016, 2020.

\bibitem{gong2018improving}
H.~Gong, Y.~Sun, Z.~Mei, and B.~Huang, ``Improving accuracy of rutting
  prediction for mechanistic-empirical pavement design guide with deep neural
  networks,'' \emph{Constr. Build. Mater.}, vol. 190, pp. 710--718, 2018.

\bibitem{yao2019establishment}
L.~Yao, Q.~Dong, J.~Jiang, and F.~Ni, ``Establishment of prediction models of
  asphalt pavement performance based on a novel data calibration method and
  neural network,'' \emph{Transp. Res. Rec.}, vol. 2673, no.~1, pp. 66--82,
  2019.

\bibitem{tang2021review}
J.~Tang, G.~Liu, and Q.~Pan, ``A review on representative swarm intelligence
  algorithms for solving optimization problems: Applications and trends,''
  \emph{IEEE-CAA J. Automatica Sin.}, vol.~8, no.~10, pp. 1627--1643, 2021.

\bibitem{wang2019parameter}
J.~Wang and T.~Kumbasar, ``Parameter optimization of interval type-2 fuzzy
  neural networks based on pso and bbbc methods,'' \emph{IEEE-CAA J. Automatica
  Sin.}, vol.~6, no.~1, pp. 247--257, 2019.

\bibitem{karat2022optimal}
C.~Karat and R.~Senthilkumar, ``Optimal resource allocation with deep
  reinforcement learning and greedy adaptive firefly algorithm in cloud
  computing,'' \emph{Concurrency and Computation: Practice and Experience},
  vol.~34, no.~4, p. e6657, 2022.

\bibitem{liu2022minimum}
H.~Liu, J.~Zhang, Q.~Liu, and J.~Cao, ``Minimum spanning tree based graph
  neural network for emotion classification using eeg,'' \emph{Neural
  Networks}, vol. 145, pp. 308--318, 2022.

\bibitem{liu2022data}
H.~Liu, J.~Cao, W.~Huang, X.~Shi, and X.~Zhou, ``A data-driven approach to the
  evaluation of asphalt pavement structures using falling weight
  deflectometer,'' \emph{Discrete Cont. Dyn-S}, 2022.

\bibitem{guo2022exponential}
L.~Guo, X.~Shi, and J.~Cao, ``Exponential convergence of primal-dual dynamical
  system for linear constrained optimization,'' \emph{IEEE-CAA J. Automatica
  Sin.}, vol.~9, no.~4, pp. 745--748, 2022.

\bibitem{islam2020holistic}
J.~Islam, P.~M. Vasant, B.~M. Negash, M.~B. Laruccia, M.~Myint, and J.~Watada,
  ``A holistic review on artificial intelligence techniques for well placement
  optimization problem,'' \emph{Adv. Eng. Softw.}, vol. 141, p. 102767, 2020.

\bibitem{ng2022well}
C.~S.~W. Ng, A.~J. Ghahfarokhi, and M.~N. Amar, ``Well production forecast in
  volve field: Application of rigorous machine learning techniques and
  metaheuristic algorithm,'' \emph{J. Pet. Sci. Eng.}, vol. 208, p. 109468,
  2022.

\bibitem{qiao2020nature}
W.~Qiao, H.~Moayedi, and L.~K. Foong, ``Nature-inspired hybrid techniques of
  iwo, da, es, ga, and ica, validated through a k-fold validation process
  predicting monthly natural gas consumption,'' \emph{Energy Build.}, vol. 217,
  p. 110023, 2020.

\bibitem{abdellatif2021optimizing}
S.~O.~E. Abdellatif, E.-Y. Mohammad, H.~A. Ghali, and W.~R. Anis, ``Optimizing
  a pv/diesel hybrid system in oil and gas industry using metaheuristic
  techniques,'' \emph{Int. J. Energy Res.}, vol.~11, no.~2, pp. 647--653, 2021.

\bibitem{xu2021structural}
X.~Xu, Y.~Gu, W.~Huang, D.~Chen, C.~Zhang, and X.~Yang, ``Structural
  optimization of steel—epoxy asphalt pavement based on orthogonal design and
  ga—bp algorithm,'' \emph{Crystals}, vol.~11, no.~4, p. 417, 2021.

\bibitem{liang2021machine}
C.~Liang, X.~Xu, H.~Chen, W.~Wang, K.~Zheng, G.~Tan, Z.~Gu, and H.~Zhang,
  ``Machine learning approach to develop a novel multi-objective optimization
  method for pavement material proportion,'' \emph{Appl. Sci.}, vol.~11, no.~2,
  p. 835, 2021.

\bibitem{ju1982control}
D.~Ju-Long, ``Control problems of grey systems,'' \emph{Syst. Control Lett.},
  vol.~1, no.~5, pp. 288--294, 1982.

\bibitem{vallee2008grey}
R.~Vall{\'e}e, ``Grey information: theory and practical applications,''
  \emph{Kybernetes}, 2008.

\bibitem{hu2018establishing}
Y.-C. Hu, Y.-J. Chiu, and J.-F. Tsai, ``Establishing grey criteria similarity
  measures for multi-criteria recommender systems.'' \emph{J. Grey Syst.},
  vol.~30, no.~1, 2018.

\bibitem{jiang2018green}
P.~Jiang, Y.-C. Hu, G.-F. Yen, and S.-J. Tsao, ``Green supplier selection for
  sustainable development of the automotive industry using grey
  decision-making,'' \emph{Sustain. Dev.}, vol.~26, no.~6, pp. 890--903, 2018.

\bibitem{blondel2008fast}
V.~D. Blondel, J.~L. Guillaume, R.~Lambiotte, and E.~Lefebvre, ``Fast unfolding
  of communities in large networks,'' \emph{J. Stat. Mech. Theory Exp.}, vol.
  2008, no.~10, p. P10008, 2008.

\bibitem{huang2004extreme}
G.~B. Huang, Q.~Y. Zhu, and C.~K. Siew, ``Extreme learning machine: a new
  learning scheme of feedforward neural networks,'' in \emph{Proceedings of the
  2004 IEEE International Joint Conference on Neural Networks}, vol.~2.\hskip
  1em plus 0.5em minus 0.4em\relax Budapest: IEEE, 2004, pp. 985--990.

\bibitem{huang2006extreme}
G.~B. \vspace{0mm}Huang, Q.~Y. Zhu, and C.~K. Siew, ``Extreme learning machine:
  theory and applications,'' \emph{Neurocomputing}, vol.~70, no. 1-3, pp.
  489--501, 2006.

\bibitem{kennedy1995particle}
J.~Kennedy and R.~Eberhart, ``Particle swarm optimization,'' in
  \emph{Proceedings of ICNN'95-International Conference on Neural Networks},
  vol.~4.\hskip 1em plus 0.5em minus 0.4em\relax Perth: IEEE, 1995, pp.
  1942--1948.

\bibitem{shi2001particle}
Y.~Shi \emph{et~al.}, ``Particle swarm optimization: developments, applications
  and resources,'' in \emph{Proceedings of the 2001 Congress on Evolutionary
  Computation}, vol.~1.\hskip 1em plus 0.5em minus 0.4em\relax Seoul: IEEE,
  2001, pp. 81--86.

\bibitem{cleveland1988locally}
W.~S. Cleveland and S.~J. Devlin, ``Locally weighted regression: an approach to
  regression analysis by local fitting,'' \emph{J. Am. Stat. Assoc.}, vol.~83,
  no. 403, pp. 596--610, 1988.

\bibitem{ester1996density}
M.~Ester, H.-P. Kriegel, J.~Sander, X.~Xu \emph{et~al.}, ``A density-based
  algorithm for discovering clusters in large spatial databases with noise.''
  in \emph{Proceedings of the 2nd International Conference Knowledge Discovery
  and Data Mining}, vol.~96, no.~34, Portland, 1996, pp. 226--231.

\bibitem{macqueen1967classification}
J.~MacQueen, ``Classification and analysis of multivariate observations,'' in
  \emph{Proceedings of the 5th Berkeley Symp. Math. Statist. Probability},
  California, 1967, pp. 281--297.

\bibitem{bishop2006pattern}
C.~M. Bishop and N.~M. Nasrabadi, \emph{Pattern recognition and machine
  learning}.\hskip 1em plus 0.5em minus 0.4em\relax Springer, 2006, vol.~4,
  no.~4.

\bibitem{ankerst1999optics}
M.~Ankerst, M.~M. Breunig, H.-P. Kriegel, and J.~Sander, ``Optics: Ordering
  points to identify the clustering structure,'' \emph{Sigmod Rec.}, vol.~28,
  no.~2, pp. 49--60, 1999.

\bibitem{wang2009cvap}
K.~Wang, B.~Wang, and L.~Peng, ``Cvap: validation for cluster analyses,''
  \emph{Data Sci. J.}, pp. 0\,904\,220\,071--0\,904\,220\,071, 2009.

\bibitem{iosifidis2015kernel}
A.~Iosifidis, A.~Tefas, and I.~Pitas, ``On the kernel extreme learning machine
  classifier,'' \emph{Pattern Recognit. Lett.}, vol.~54, pp. 11--17, 2015.

\bibitem{hochreiter1997long}
S.~Hochreiter and J.~Schmidhuber, ``Long short-term memory,'' \emph{Neural
  Comput.}, vol.~9, no.~8, pp. 1735--1780, 1997.

\bibitem{ghritlahre2018exergetic}
H.~K. Ghritlahre and R.~K. Prasad, ``Exergetic performance prediction of solar
  air heater using mlp, grnn and rbf models of artificial neural network
  technique,'' \emph{J. Environ. Manage.}, vol. 223, pp. 566--575, 2018.

\bibitem{cho2014learning}
K.~Cho, B.~Van~Merri{\"e}nboer, C.~Gulcehre, D.~Bahdanau, F.~Bougares,
  H.~Schwenk, and Y.~Bengio, ``Learning phrase representations using rnn
  encoder-decoder for statistical machine translation,'' \emph{arXiv preprint
  arXiv:1406.1078}, 2014.

\bibitem{young1977proof}
I.~T. Young, ``Proof without prejudice: use of the kolmogorov-smirnov test for
  the analysis of histograms from flow systems and other sources.''
  \emph{Journal of Histochemistry \& Cytochemistry}, vol.~25, no.~7, pp.
  935--941, 1977.

\bibitem{shi1999empirical}
Y.~Shi and R.~C. Eberhart, ``Empirical study of particle swarm optimization,''
  in \emph{Proceedings of the 1999 Congress on Evolutionary Computation-CEC99},
  vol.~3.\hskip 1em plus 0.5em minus 0.4em\relax Washington: IEEE, 1999, pp.
  1945--1950.

\bibitem{Li2022CPSO}
Z.~Li, X.~Shi, J.~Cao, X.~Wang, and W.~Huang, ``{CPSO-XGBoost} segmented
  regression model for asphalt pavement deflection basin area prediction,''
  \emph{Sci. China-Technol. Sci.}, vol.~65, no.~7, pp. 1470--1481, 2022.

\bibitem{whitley1994genetic}
D.~Whitley, ``A genetic algorithm tutorial,'' \emph{Stat. Comput.}, vol.~4,
  no.~2, pp. 65--85, 1994.

\bibitem{wang2012annual}
J.~Wang, L.~Li, D.~Niu, and Z.~Tan, ``An annual load forecasting model based on
  support vector regression with differential evolution algorithm,''
  \emph{Appl. Energy}, vol.~94, pp. 65--70, 2012.

\bibitem{mirjalili2016whale}
S.~Mirjalili and A.~Lewis, ``The whale optimization algorithm,'' \emph{Adv.
  Eng. Softw.}, vol.~95, pp. 51--67, 2016.

\bibitem{chen2016xgboost}
T.~Chen and C.~Guestrin, ``Xgboost: A scalable tree boosting system,'' in
  \emph{Proceedings of the 22nd acm sigkdd international conference on
  knowledge discovery and data mining}.\hskip 1em plus 0.5em minus 0.4em\relax
  San Francisco: ACM, 2016, pp. 785--794.

\bibitem{smola2004tutorial}
A.~J. Smola and B.~Sch{\"o}lkopf, ``A tutorial on support vector regression,''
  \emph{Stat. Comput.}, vol.~14, no.~3, pp. 199--222, 2004.

\bibitem{yang2019evaluation}
Q.~Yang and Y.~Deng, ``Evaluation of cracking in asphalt pavement with
  stabilized base course based on statistical pattern recognition,'' \emph{Int.
  J. Pavement Eng.}, vol.~20, no.~4, pp. 417--424, 2019.

\bibitem{smith2013comparison}
P.~F. Smith, S.~Ganesh, and P.~Liu, ``A comparison of random forest regression
  and multiple linear regression for prediction in neuroscience,'' \emph{J.
  Neurosci. Methods}, vol. 220, no.~1, pp. 85--91, 2013.

\bibitem{kurt2018classification}
S.~Kurt, E.~{\"O}z, {\"O}.~E. A{\c{s}}k{\i}n, and Y.~Y. {\"O}z,
  ``Classification of nucleotide sequences for quality assessment using
  logistic regression and decision tree approaches,'' \emph{Neural Comput.
  Appl.}, vol.~29, no.~8, pp. 251--262, 2018.

\bibitem{pedregosa2011scikit}
F.~Pedregosa, G.~Varoquaux, A.~Gramfort, V.~Michel, B.~Thirion, O.~Grisel,
  M.~Blondel, P.~Prettenhofer, R.~Weiss, V.~Dubourg \emph{et~al.},
  ``Scikit-learn: Machine learning in python,'' \emph{J. Mach. Learn. Res.},
  vol.~12, pp. 2825--2830, 2011.

\end{thebibliography}

%

\begin{IEEEbiography}[{\includegraphics[width=1in,height=1in,clip,keepaspectratio]{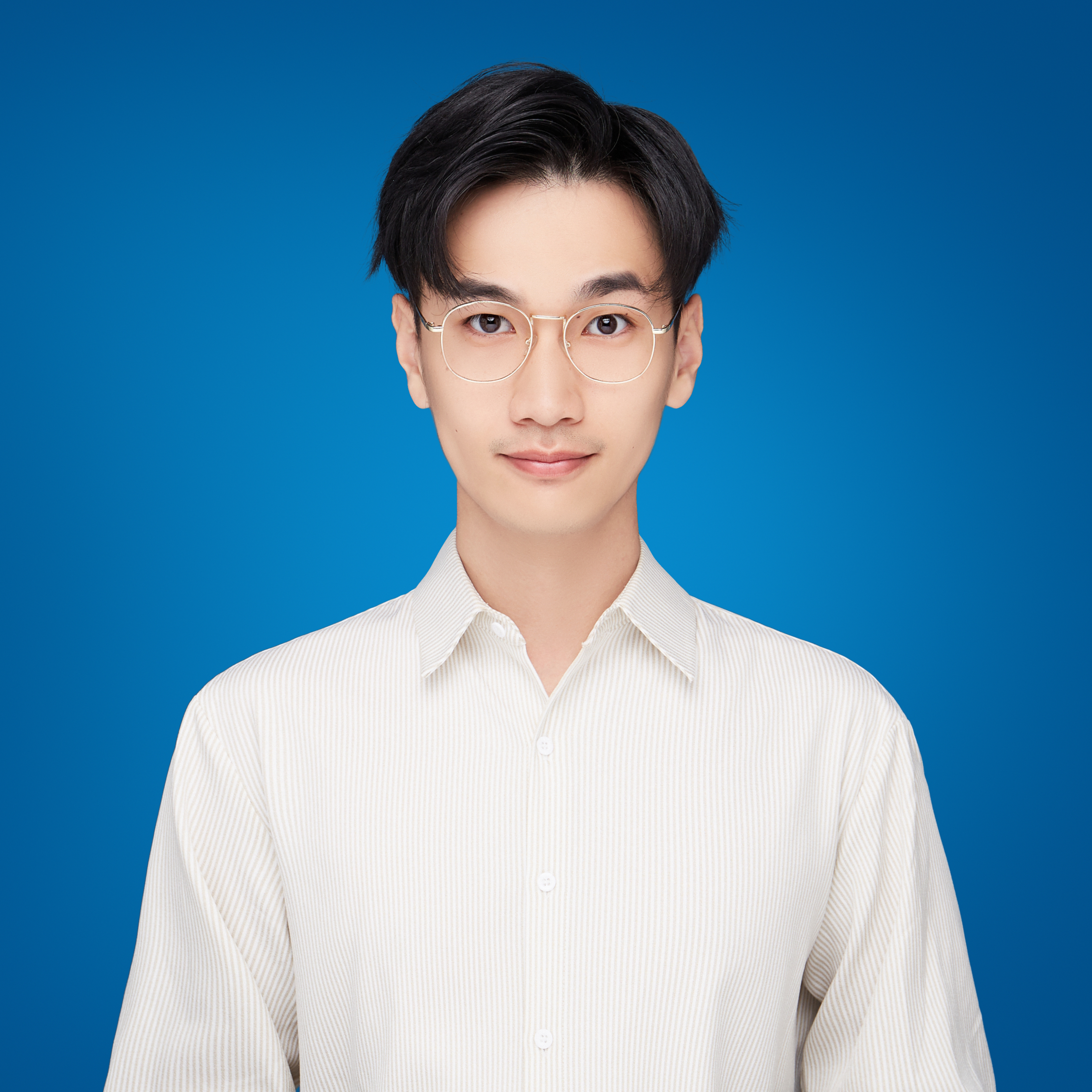}}]{Zhuoxuan Li}
received the B.S. degree from Shenyang University of Technology, Shenyang, China, in 2019. He received the M.S. degree in applied mathematics from Southeast University, Nanjing, China, in 2022. He is currently pursuing the Ph.D. degrees with the School of  Mathematics, Southeast University, Nanjing, China. His current research interests data mining, man-machine game, machine learning, intelligent algorithms and neural networks.
\end{IEEEbiography}

\begin{IEEEbiography}[{\includegraphics[width=1in,height=1.25in,clip,keepaspectratio]{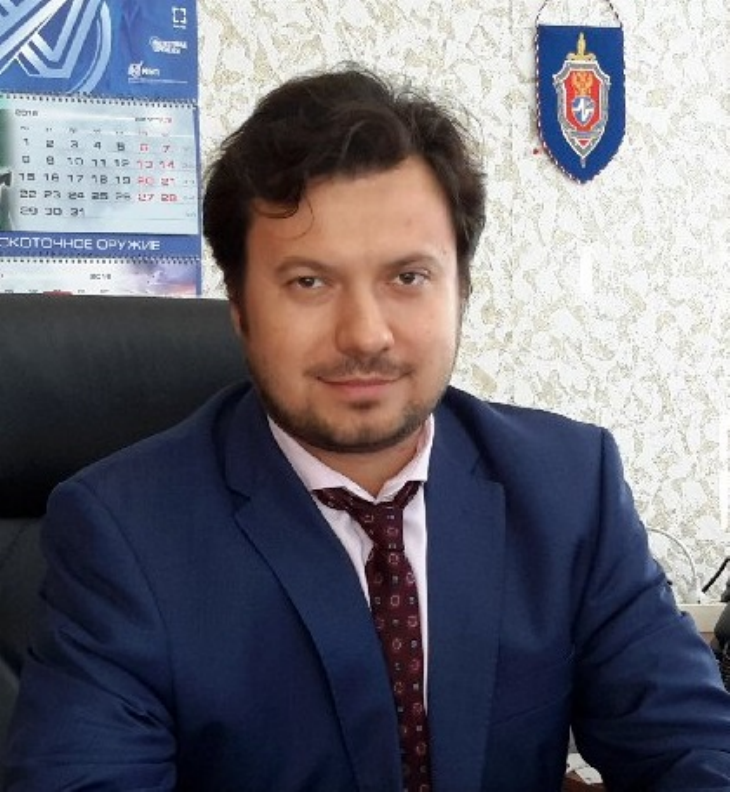}}]{Iakov Korovin}
graduated from Taganrog Radio Engineering University,  Taganrog, Russia with a M.S. degree in Information Systems in Economics. He received the Ph.D degree in Engineering in 2009,  worked as the head of the laboratory of neural network systems of the Scientific Research Institute of Multiprocessor Computer Systems, Southern Federal University. He is currently a Principal of this Institute. His research interests include Data Mining methods, neural network technologies, decision support systems, image processing.
\end{IEEEbiography}

\begin{IEEEbiography}[{\includegraphics[width=1in,height=1.25in,clip,keepaspectratio]{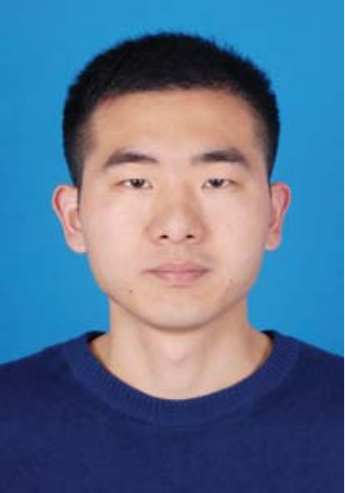}}]{Xinli Shi}
received the B.S. degree in software engineering, the M.S. degree in applied mathematics, and the Ph.D. degree in control science and engineering from Southeast University, Nanjing, China, in 2013, 2016, and 2019, respectively. Dr. Shi was a recipient of the Outstanding Ph. D.  Degree Thesis Award from Jiangsu Province, China, in 2020. He is currently an associate professor with the School of Cyber Science and Engineering, Southeast University, Nanjing, China.

His current research interests include model
predictive control, distributed optimization, smart grids, and network control systems.
\end{IEEEbiography}

\begin{IEEEbiography}[{\includegraphics[width=1in,height=1.25in,clip,keepaspectratio]{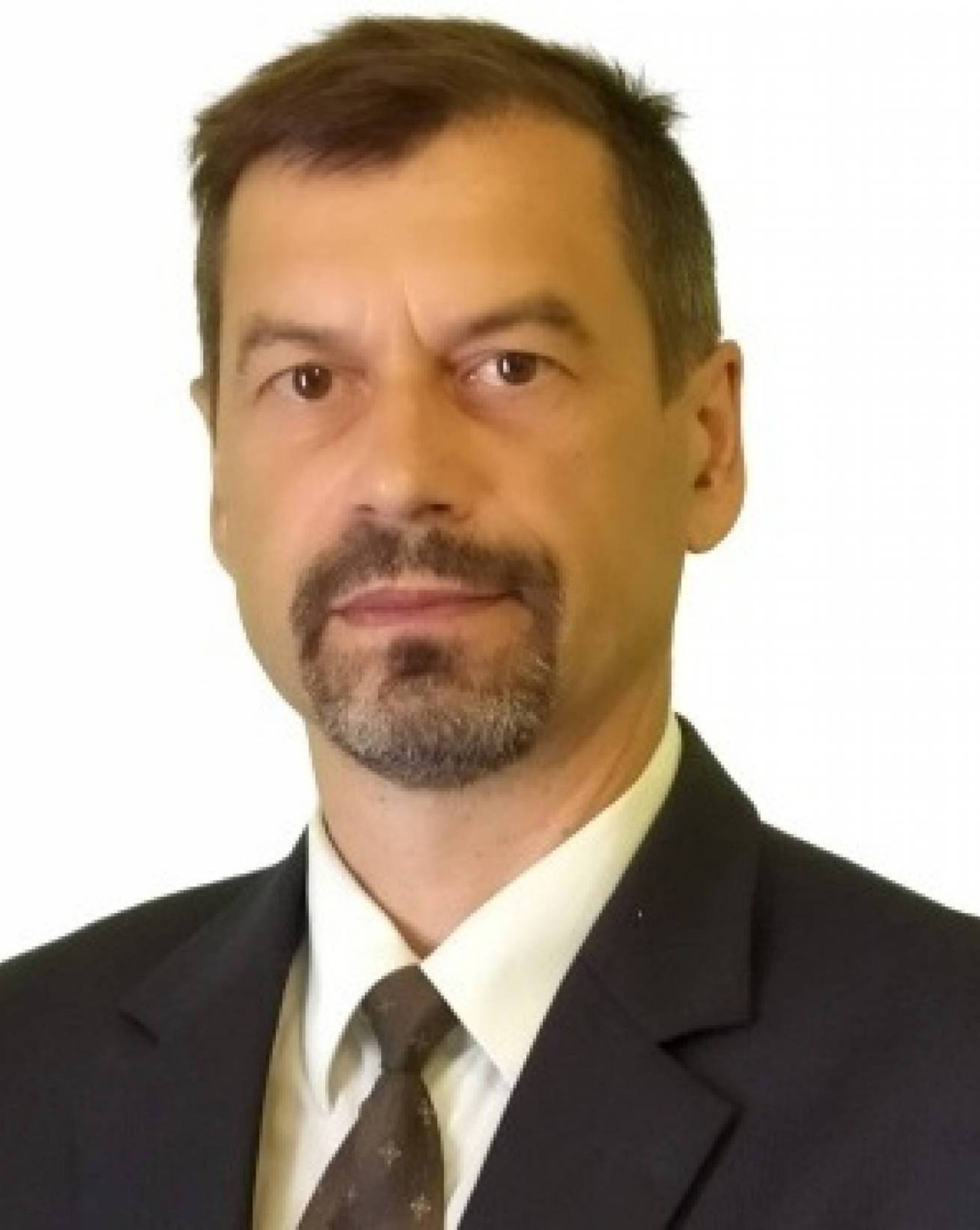}}]{Sergey Gorbachev}
received High Diploma with Honors in applied mathematics in 1993 and Ph.D degree in mathematical modeling, 
computational methods and software complex in 2003, all in Department of Applied Mathematics and Cybernetics of the National research Tomsk state University, Tomsk, Russia. He worked as a Senior Researcher at the International Laboratory of Technical Vision Systems. He has authored or edited 28 books, 20 patents and copyright certificates and more than 130 research publications in refereed journals and conference proceedings. He is a full member of the Russian Academy of Engineering. His research interests include neural-fuzzy networks, ensemble models, soft computing, deep learning, pattern recognition, image processing, Big Data, 
data mining and neutrosophic cognitive maps.
\end{IEEEbiography}

\begin{IEEEbiography}[{\includegraphics[width=1in,height=1.25in,clip,keepaspectratio]{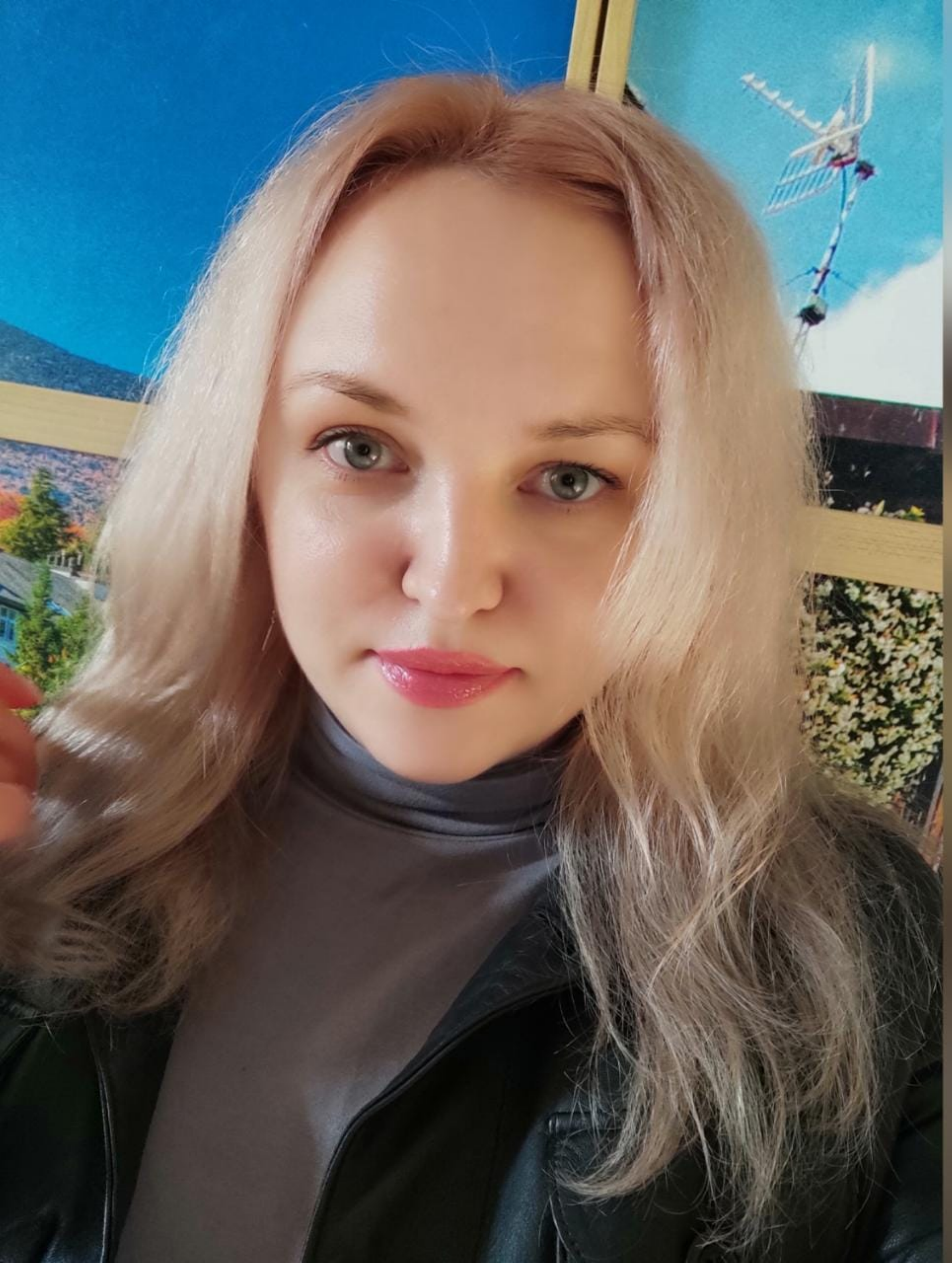}}]{Nadezhda Gorbacheva}
received High Diploma in technologies in 2009 in National research Tomsk polytechnical University, Tomsk, Russia. Her research interests include neural networks, hybrid methods, BiG Data analysis, soft computing, pattern recognition.
\end{IEEEbiography}

\begin{IEEEbiography}
[{\includegraphics[width=1in,height=1.25in,clip,keepaspectratio]{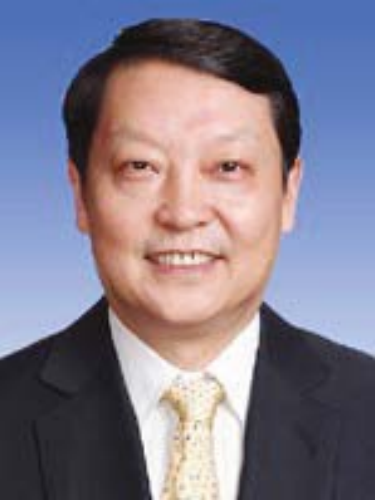}}]{Wei Huang}
received the B.S., M.S. and Ph.D. degrees in road engineering from Southeast University, in 1982, 1986 and 1995, respectively.

He is a distinguished professor in Civil Engineer at the Intelligent Transportation System Research Center of the Southeast University, Nanjing, Jiangsu Province, P.R. China. He has published 13 books.

Dr. Huang was a recipient of 26 awards from both the National and Provincial Level. He enjoys the State Council special allowance and receives supports from the New Century Talent Program, the National Outstanding Mid-Aged Experts Program, the National Talents
Engineering Program, and the Yangtze Scholar Program from various agencies and organizations. He is one of the forerunners in the research fields of long span steel bridge pavement and intelligent transportation systems of China. He is a member of Chinese Academy of Engineering.
\end{IEEEbiography}

\begin{IEEEbiography}
[{\includegraphics[width=1in,height=1.25in,clip,keepaspectratio]{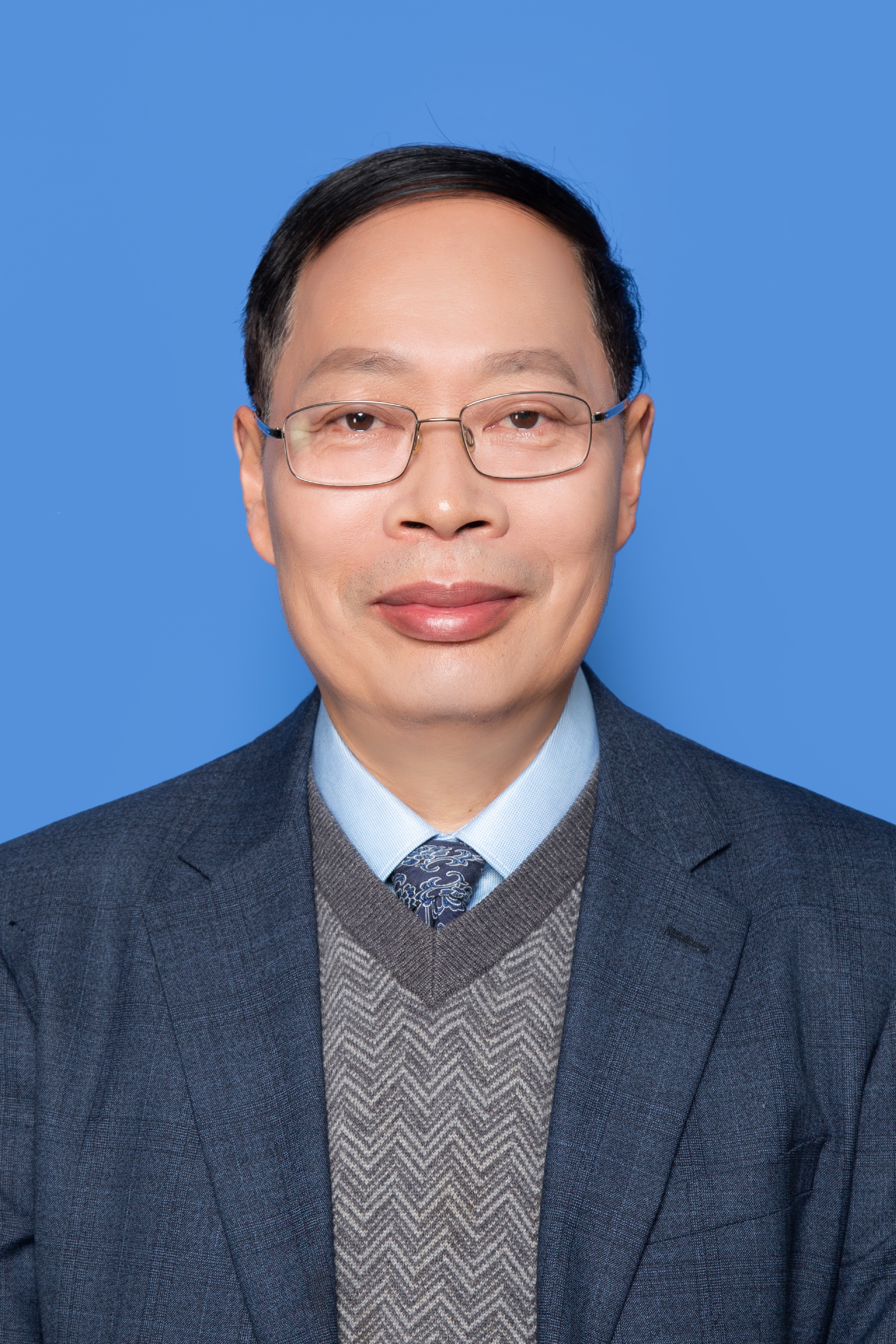}}]{Jinde Cao}
(Fellow, IEEE) received the B.S. degree from Anhui Normal University, Wuhu, China, the M.S. degree from Yunnan University, Kunming, China, and the Ph.D. degree from Sichuan University, Chengdu, China, all in mathematics/applied mathematics, in 1986, 1989, and 1998, respectively.

He is an Endowed Chair Professor, the Dean of the School of Mathematics, the Director of the Jiangsu Provincial Key Laboratory of Networked Collective Intelligence of China and the Director of the Research Center for Complex Systems and Network Sciences at Southeast University. Prof. Cao was a recipient of the National Innovation Award of China, Gold medal of Russian Academy of Natural Sciences, Obada Prize and the Highly Cited Researcher Award in Engineering, Computer Science, and Mathematics by Thomson Reuters/Clarivate Analytics. He is elected as a member of the Academy of Europe, a foreign member of Russian Academy of Sciences, a foreign member of Russian Academy of Engineering, a member of the European Academy of Sciences and Arts, a foreign fellow of Pakistan Academy of Sciences, a fellow of African Academy of Sciences, a foreign Member of the Lithuanian Academy of Sciences, and an IASCYS academician.
\end{IEEEbiography}




\end{document}